\documentclass{article}

\usepackage[preprint]{neurips_2021}

\usepackage[utf8]{inputenc} 
\usepackage[T1]{fontenc}    
\usepackage{hyperref}       
\usepackage{url}            
\usepackage{booktabs}       
\usepackage{amsfonts}       
\usepackage{nicefrac}       
\usepackage{microtype}      
\usepackage{xcolor}         
\usepackage{subcaption}
\usepackage{amssymb,amsmath}
\usepackage{bm}
\usepackage{adjustbox}
\usepackage{wrapfig}
\usepackage[capitalise]{cleveref}

\setcitestyle{numbers,square}
\hypersetup{colorlinks=true,citecolor=blue}

\title{Nonlinear Hawkes Processes in Time-Varying System}

\author{%
Feng Zhou$^{1}$, Quyu Kong$^{2}$, Yixuan Zhang$^{3}$, Cheng Feng$^{4}$, Jun Zhu$^{1}$\\
$^{1}$Dept. of Comp. Sci. \& Tech., BNRist Center, THU-Bosch Joint ML Center, Tsinghua University\\
$^{2}$Data Science Institute, University of Technology Sydney\\
$^{3}$Research School of Computer Science, Australian National University\\
$^{4}$Siemens Technology\\
\texttt{\{zhoufeng6288, dcszj\}@tsinghua.edu.cn},
\texttt{quyu.kong@anu.edu.au},\\
\texttt{yixuan.zhang@uts.edu.au},
\texttt{cheng.feng@siemens.com}
}

\begin{document}

\maketitle

\begin{abstract}
Hawkes processes are a class of point processes that have the ability to model the self- and mutual-exciting phenomena. Although the classic Hawkes processes cover a wide range of applications, their expressive ability is limited due to three key hypotheses: parametric, linear and homogeneous. Recent work has attempted to address these limitations separately. This work aims to overcome all three assumptions simultaneously by proposing the flexible state-switching Hawkes processes: a flexible, nonlinear and nonhomogeneous variant where a state process is incorporated to interact with the point processes. The proposed model empowers Hawkes processes to be applied to time-varying systems. For inference, we utilize the latent variable augmentation technique to design two efficient Bayesian inference algorithms: Gibbs sampler and mean-field variational inference, with analytical iterative updates to estimate the posterior. In experiments, our model achieves superior performance compared to the state-of-the-art competitors. 
\end{abstract}

\section{Introduction}
\label{intro}
Hawkes processes (HPs)~\citep{hawkes1971spectra} are a class of point processes that can model the self- and mutual-exciting phenomena in many research disciplines where the occurrence of events in the past intensifies the rate of new events in the future. Given the ability to capture clustering and contagion effects, the classic HPs have been applied in a wide range of domains including high-frequency financial trade~\citep{bacry2015hawkes}, neural spike trains~\citep{paninski2004maximum,zhou2020efficient}, seismology~\citep{ogata1998space,ogata1999seismicity} and transportation~\citep{du2016recurrent,zhou2018refined}. 

HPs have a more powerful expressive ability than Poisson processes~\citep{kingman2005p} due to the relief of the complete independence assumption~\citep{daley2007introduction}. However, the vanilla HPs still have three key limitations: \textbf{(1)} \textit{Parametric}, the influence functions within and across dimensions are assumed to be specific parametric functions, e.g., exponential decay or power law decay, which eases the inference at the cost of model expressiveness. In real-world problems, actual influence functions can be rather complex and vary a lot among different applications~\citep{mohler2011self,zhou2013learning}. \textbf{(2)} \textit{Linear}, the conditional intensity (\cref{eq1}) of HPs is a linear superposition of influence functions, so the influence functions must be non-negative (excitatory) to guarantee a non-negative conditional intensity. The linear assumption leads to the failure of incorporating negative (inhibitive) effects into HPs. The inhibitive effect is an important characteristic in some domains, e.g., a spike fired by the pre-synaptic neuron may inhibit the post-synaptic neuron from firing its own spikes in neuroscience~\citep{mongillo2018inhibitory,zhou2020efficient}. \textbf{(3)} \textit{Homogeneous}, the underlying parameters of HPs, i.e., the base intensities and influence functions in \cref{eq1}, are time-invariant. The homogeneous assumption is inconsistent with real-world applications where event dynamics are temporally heterogeneous and depend on the state of the system. For example, in the high-frequency financial data, the arrival rate of orders is influenced by the state of the limit order book, such as the bid-ask spread or the volume imbalance~\citep{morariu2018state}.

\paragraph{Motivation}
Much existing work has been done to address one or two of those issues while leaving the rest unaddressed. \Cref{tab1} provides a non-exhaustive list with details provided in \cref{relatedwork}. To the best of our knowledge, no model has been proposed yet to address all the three limitations simultaneously. To circumvent all issues together, we propose the flexible state-switching HPs (FS-Hawkes). The motivation is to empower HPs to model time-varying point processes with flexible and exciting-inhibitive influence functions. Our proposed model permits the latent interaction structure of HPs to change with time, e.g., the presynaptic neuron excites the postsynaptic one in one state but inhibits it in another state, which previous models are incapable of inferring. 

\begin{wraptable}{r}{0.5\textwidth}\vspace{-.4cm}
\caption{The comparison of our work with some existing works w.r.t. nonparameric, nonlinear and nonhomogeneous.}
\label{tab1}
\begin{center}
\scalebox{0.59}{
\begin{tabular}{ccccc}
\toprule
Work & Nonparametric\footnotemark & Nonlinear & Nonhomogeneous\\
\midrule
\citet{marsan2008extending} & \checkmark & $\times$ & $\times$\\
\citet{lewis2011nonparametric} & \checkmark & $\times$ & $\times$\\
\citet{zhou2013learning} & \checkmark & $\times$ & $\times$\\
\citet{bacry2016first} & \checkmark & $\times$ & $\times$\\
\citet{zhang2018efficient} & \checkmark & $\times$ & $\times$\\
\citet{zhou2019efficient} & \checkmark & $\times$ & $\times$\\
\citet{zhou2020auxiliary} & \checkmark & $\times$ & $\times$\\
\citet{gerhard2017stability} & $\times$ & \checkmark & $\times$\\
\citet{apostolopoulou2019mutually} & $\times$ & \checkmark & $\times$\\
\citet{zhou2020efficient} & \checkmark & \checkmark & $\times$\\
\citet{wang2012markov} & $\times$ & $\times$ & \checkmark\\
\citet{wu2019markov} & $\times$ & $\times$ & \checkmark\\
\citet{zhou2020fast} & \checkmark & $\times$ & \checkmark\\
\citet{morariu2018state} & $\times$ & $\times$ & \checkmark\\
our work & \checkmark & \checkmark & \checkmark\\
\bottomrule
\end{tabular}}
\end{center}
\vskip -0.2in
\end{wraptable}
\footnotetext{The influence function estimation in some works is flexible not nonparameric. To be concise, we still list them in the nonparametric column.}
\paragraph{Contribution}
Our contributions are: \textbf{(1)} we propose a novel \textit{flexible, nonlinear} and \textit{nonhomogeneous} HPs variant that has flexible influence patterns, is able to handle inhibitive effects, and has state process driven (time-varying) parameters simultaneously; and \textbf{(2)} we develop two efficient Bayesian inference algorithms, a \textit{Gibbs sampler} and a \textit{mean-field variational inference} method, that leverage latent variable augmentation techniques~\citep{chen2013scalable,linderman2017recurrent,polson2013bayesian} to obtain closed-form iterative updates. It is worth noting that, although some work also used the state process to describe time-varying parameters, it cannot simultaneously address the three limitations as our model does. Besides, its inference methods have no closed-form solutions and are inefficient compared to our methods.

\section{Our Model}
\label{model}

Our model has three key components: flexible influence functions, inhibitive effects and a coupled state process, corresponding to the relief of parametric, linear and homogeneous assumptions, respectively. In this section, we review how the vanilla multivariate HPs are extended to include these components. 

\subsection{Vanilla Multivariate Hawkes Processes}

An $M$-dimensional multivariate HP~\citep{hawkes1971spectra} consists of $M$ sequences of random timestamps $\{\{t_n^i \}_{n=1}^{N_i}\}_{i=1}^M$ in the observation window $[0,T]$ where $t_n^i$ is the time of $n$-th event on $i$-th dimension, $M$ is the number of dimensions and $N_i$ is the number of events in the $i$-th dimension. A point process can be identified by its counting process $N(t)=\sum_{n}I(t_n<t)$ where $I(\cdot)$ is the indicator function. The conditional intensity $\lambda(t\mid\mathcal{H}_t)$ is defined as $\lambda(t\mid\mathcal{H}_t)dt=p(N(t+dt)-N(t)=1\mid\mathcal{H}_t)$ where $\mathcal{H}_t$ is the historical information before $t$. In the sequel, we use $\lambda^*(t)$ to represent the conditional intensity $\lambda(t\mid\mathcal{H}_t)$ for short. The $i$-th dimensional conditional intensity is designed as
\begin{equation}
\label{eq1}
\lambda_i^*(t)=\mu_i+\sum_{j=1}^M\sum_{t_n^j<t}f_{ij}(t-t_n^j),
\end{equation}
where $\mu_i$ is the base intensity and $f_{ij}(\cdot)$ is the causal influence function from dimension $j$ to dimension $i$. To ease inference, $f_{ij}(\cdot)$ is typically assumed to be exponential decay or power law decay. The summation over past events and dimensions leads to the self- and mutual-excitation as the occurrence of past events within and across dimensions increases the future intensity. 

\subsection{Flexible State-Switching Hawkes Processes}
\label{sec2.2}

\begin{wrapfigure}{r}{0.4\textwidth}\vspace{-.6cm}
\begin{center}
\includegraphics[width=0.4\textwidth]{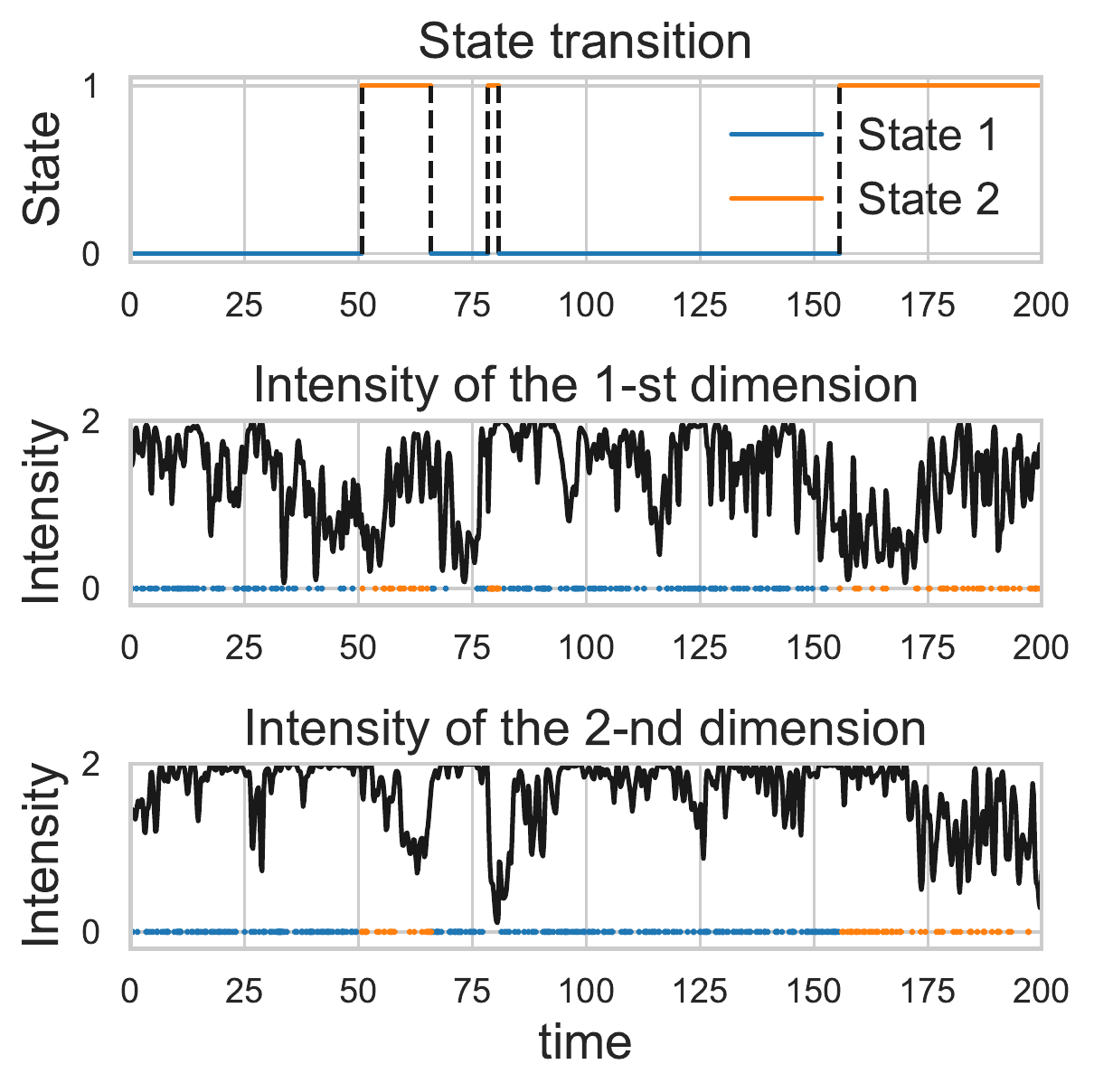}
\caption{A realization of the 2-state 2-dimensional FS-Hawkes on $[0,200]$. The 1-st plot shows the evolution of the state process; the 2-nd and 3-rd plots show the generated 2-dimensional timestamps and conditional intensities. The point process is generated by two sets of parameters $\bm{\theta}_1$ (in state $1$) and $\bm{\theta}_2$ (in state $2$). The point's color indicates that the timestamps are generated by the corresponding parameters.}
\label{fig1}
\end{center}
\vskip -0.3in
\end{wrapfigure}

As mentioned, three assumptions (parameterization, linearity and homogeneity) restrict the expressiveness of HPs. We now propose the FS-Hawkes to enrich the vanilla model with flexible influence functions, inhibitory effects and time-varying parameters. An $M$-dimensional FS-Hawkes consists of $M$ sequences of random timestamps and their corresponding states $D=\{\{\{t_n^i, z(t_n^i)\}_{n=1}^{N_i}\}_{i=1}^M, z(T)\}$ in the observation window $[0,T]$. A schematic example of a 2-state 2-dimensional FS-Hawkes is shown in \cref{fig1}. In this example, the process has two sets of parameters corresponding to two states. 

\paragraph{The State Process}
In the FS-Hawkes, we introduce a state process $z(t)$ that takes values in a discrete finite state space $\mathcal{Z}=\{1,\ldots,K\}$ to represent the system state. Similar to \cite{morariu2018state}, we establish an endogenous Markov state process which is coupled with the point processes to form a closed-loop interaction. Reciprocally, the underlying parameters of point processes depend on the current system state; at the meantime, the state process switches only when an event occurs on point processes by a state-transition matrix depending on the event type. Given a set of state-transition matrices $\bm{\Phi}=\{\bm{\Phi}_{1},\ldots,\bm{\Phi}_{M}\}$ with $\bm{\Phi}_{i}$ being a $K\times K$ transition probability matrix for the $i$-th dimension (type), the transition probability of $z(t)$ at event timestamp $t_n^i$ is
\begin{equation}
p(z({t_n^i}^+)=k'\mid z(t_n^i)=k)=\phi_{i}(k,k'),
\label{eq2}
\end{equation}
where we assume the state process $z(t)$ is left continuous $\lim_{t\to c^-}z(t)= z(c)$ and $z({t_n^i}^+)$ is the right limit of $z({t_n^i})$. $\phi_{i}(k,k')$ with $k,k'\in\{1,\ldots,K\}$ is the entry of $\bm{\Phi}_{i}$. 

\paragraph{The Point Processes}
\Cref{eq2} describes how the state process evolves with the point processes. In turn, we define how the point processes depend on the state process. 
Inspired by \cite{zhou2020efficient}, we establish a nonlinear HP variant with the excitation-inhibition-mixture influence functions depending on the system state. The $i$-th dimensional conditional intensity of FS-Hawkes is defined as 
\begin{equation}
\lambda_i^*(t,z(t))=\overline{\lambda}_i\sigma(h_i(t,z(t))),\ \ \ h_i(t,z(t))=\mu_i^{z(t)}+\sum_{j=1}^M\sum_{t_n^j<t}f_{ij}^{z(t)}(t-t_n^j),
\label{eq3}
\end{equation}
where $h_i(t,z(t))$ is a real-valued state-dependent activation passed through a logistic function $\sigma(\cdot)$ to guarantee the non-negativity of intensity and then scaled by an upper-bound $\overline{\lambda}_i$. $\mu_i^{z(t)}$ and $f_{ij}^{z(t)}$ are the $z(t)$-state base activation and influence function which can be negative. The logistic nonlinearity is chosen because the P\'{o}lya-Gamma augmentation~\citep{polson2013bayesian} can be used to make inference easy and fast. 

To be flexible, $f_{ij}^{z(t)}$ is assumed to be a mixture function $f_{ij}^{z(t)}(\cdot)=\sum_{b=1}^B w_{ijb}^{z(t)}\widetilde{f}_b(\cdot)$ where $\{\widetilde{f}_b\}_{b=1}^B$ are predefined basis functions and $w_{ijb}^{z(t)}$ is the state-dependent mixture weight characterizing the influence from $j$-th dimension to $i$-th dimension by $b$-th basis function in the $z(t)$ state. Therefore, the $i$-th dimensional activation can be rewritten in a vector form
\begin{equation}
\begin{aligned}
&h_i(t,z(t))
=\mu_i^{z(t)}+\sum_{j=1}^M\sum_{b=1}^B w_{ijb}^{z(t)}F_{jb}(t)={\mathbf{w}_{i}^{z(t)}}^\top\cdot\mathbf{F}(t),
\end{aligned}
\label{eq4}
\end{equation}
where $F_{jb}(t)=\sum_{t_n^j<t}\widetilde{f}_{b}(t-t_n^j)$ is $j$-th dimensional cumulative influence on $t$ by $b$-th basis function and can be precomputed; $\mathbf{w}_{i}^{z(t)}=[\mu_i^{z(t)}, w_{i11}^{z(t)},\dotsc, w_{iMB}^{z(t)}]^\top$ and $\mathbf{F}(t)=[1, F_{11}(t), \dotsc, F_{MB}(t)]^\top$. 

Each basis function characterizes one component of the influence function that captures the temporal dynamics. Although the basis functions can be arbitrarily chosen, in our model setting, they are assumed to be the scaled shifted Beta densities on the support $[0,T_f]$ as the inference will be free from edge effects due to the finite support~\citep{kottas2006dirichlet}.

Combining \cref{eq2,eq3,eq4}, we obtain the FS-Hawkes. The FS-Hawkes successfully addresses three limitations mentioned above, as its influence functions are flexible, its influence effect can be inhibitive and its underlying parameters are time-varying. The FS-Hawkes can be considered as a closed-loop interactive system in which the parameters comprise (1) the dimension-dependent state-transition matrices, (2) the intensity upper-bounds and (3) the state-dependent activation weights, which we write as $\bm{\theta}=\{\{\bm{\Phi}_i\}_{i=1}^M,\{\overline{\lambda}_i\}_{i=1}^M,\{\mathbf{w}_{i}^k\}_{i,k=1}^{M,K}\}$. 

\section{Bayesian Inference}
\label{inference}

In this section, we present an efficient Gibbs sampler and a mean-field variational inference algorithm for Bayesian inference on FS-Hawkes, by using latent variable augmentation techniques. Specifically, given a realization $D$, the likelihood of FS-Hawkes is \begin{equation}
\label{eq5}
\begin{aligned}
p(D\mid\{\bm{\Phi}_i,\overline{\lambda}_i,\mathbf{w}_{i}^k\})=\prod_{i=1}^M\prod_{n=1}^{N_i}\phi_{i}(z(t_n^i),z(t_n^{i+}))\overline{\lambda}_i\sigma(h_i(t_n^i,z(t_n^i)))\exp\left(-\int_0^T\overline{\lambda}_i\sigma(h_i(t,z(t)))dt\right).
\end{aligned}
\end{equation}
Following the Bayesian framework, 
we place a conjugate Dirichlet prior on each row of the transition matrix $\bm{\Phi}_i$, an improper prior~\citep{bishop2006pattern} on $\overline{\lambda}_i$ and a symmetric Gaussian prior on $\mathbf{w}_{i}^k$, writing
\begin{equation}
\label{eq6}
\begin{gathered}
p(\bm{\phi}_k^i\mid\bm{\alpha})=\text{Dir}(\bm{\phi}_k^i\mid\bm{\alpha}),\ \ \ p(\overline{\lambda}_i)\propto1/\overline{\lambda}_i,\ \ \ p(\mathbf{w}_{i}^k\mid\mathbf{K})=\mathcal{N}(\mathbf{w}_{i}^k\mid\mathbf{0},\mathbf{K}),
\end{gathered}
\end{equation}
where $\bm{\phi}_k^i$ is $k$-th row of $\bm{\Phi}_i$. The hyperparameters are $\bm{\alpha}=[\alpha_1,\ldots,\alpha_K]^\top$ and covariance $\mathbf{K}$. 

Combining \cref{eq5,eq6}, we obtain the joint density over all variables. The posterior of the transition matrix is easy to compute because the Dirichlet prior is conjugate to the state process likelihood (categorical distribution). 
However, the non-conjugacy between the point process likelihood and priors renders the inference challenging. Many methods are proposed to circumvent the non-conjugate problem, such as Laplace approximation~\citep{tierney1986accurate} and expectation propagation~\citep{minka2001expectation}, but
here we leverage auxiliary latent variables to augment the likelihood in such a way that the augmented likelihood becomes conjugate to the Gaussian prior. Based on the augmented model, we construct an efficient Gibbs sampler which accurately characterizes the posterior; to further improve efficiency, a mean-field variational inference is developed to provide an approximated posterior. 

\subsection{Augmented Likelihood}
Two classes of auxiliary latent variables, the latent P\'{o}lya-Gamma variables and the latent marked Poisson processes, are augmented to convert the non-conjugate likelihood to a conjugate one. 

\paragraph{P\'{o}lya-Gamma Variables}
\citet{polson2013bayesian} introduced a Gaussian representation of the logistic function in terms of P\'{o}lya-Gamma variables 
\begin{equation}
\label{eq7}
\sigma(x)=\frac{e^x}{1+e^x}=\int_0^\infty e^{g(\omega,x)}\text{PG}(\omega\mid1,0)d\omega,
\end{equation}
where $\text{PG}(\omega\mid 1,0)$ is the P\'{o}lya-Gamma distribution and $g(\omega,x)=x/2-x^2\omega/2-\log2$. By substituting \cref{eq7} into \cref{eq5}, the products of $\sigma(h_i(t_n^i,z(t_n^i)))$ are transformed into a Gaussian form w.r.t. $\mathbf{w}_{i}^k$ because $h_i(t,k)$ is linear in $\mathbf{w}_{i}^k$. 

\paragraph{Marked Poisson Processes}
As shown in \cite{donner2018efficient,zhou2020auxiliary,zhou2020efficient}, a latent marked Poisson process can be augmented to render the exponential integral term in \cref{eq5} appear in a Gaussian form w.r.t. $\mathbf{w}_{i}^k$. 
Utilizing \cref{eq7} with the logistic symmetry property $\sigma(x)=1-\sigma(-x)$ and 
Campbell's theorem \citep{kingman2005p}, the exponential integral term can be rewritten as 
\begin{equation}
\label{eq8}
\begin{aligned}
&\exp(-\int_0^T\overline{\lambda}_i\sigma(h_i(t,z(t)))dt)=\mathbb{E}_{p_{\lambda_i}}\prod_{\Pi_i}e^{g(\omega,-h_i(t,z(t)))},
\end{aligned}
\end{equation}
where $\Pi_i=\{(\omega_r^i,t_r^i)\}_{r=1}^{R_i}$ is a random realization of a marked Poisson process and $p_{\lambda_i}$ is the probability measure of $\Pi_i$ with intensity $\lambda_i(t,\omega)=\overline{\lambda}_i\text{PG}(\omega\mid1,0)$. The timestamps $\{t_r^i\}_{r=1}^{R_i}$ follow a homogeneous Poisson process with a constant intensity $\overline{\lambda}_i$ and the latent P\'{o}lya-Gamma variable $\omega_r^i\sim\text{PG}(\omega\mid1,0)$ is the i.i.d. mark on each timestamp $t_r^i$. The detailed derivation of \cref{eq8} is provided in the appendix. After augmenting the latent marked Poisson processes, the exponential integral term appears in a Gaussian form w.r.t. $\mathbf{w}_{i}^k$. 

Substituting \cref{eq7,eq8} into the likelihood \cref{eq5}, we obtain the augmented likelihood which contains $\mathbf{w}_{i}^k$ in a Gaussian form and is thus conjugate to the Gaussian prior,
\begin{equation}
\begin{aligned}
&p(D,\{\bm{\omega}_i\},\{\Pi_i\}\mid\{\bm{\Phi}_i\},\{\overline{\lambda}_i\},\{\mathbf{w}_{i}^k\})=\\
&\prod_{i=1}^M\left[\prod_{n=1}^{N_i}\left[\lambda_{i}(t_n^i,\omega_n^i)e^{g(\omega_n^i,h_i(t_n^i,z(t_n^i)))}\right]p_{\lambda_i}(\Pi_i\mid\overline{\lambda}_i)\prod_{\Pi_i}e^{g(\omega,-h_i(t,z(t)))}\right]\prod_{i=1}^M\prod_{n=1}^{N_i}\phi_{i}(z(t_n^i),z(t_n^{i+})), 
\end{aligned}
\label{eq9}
\end{equation}
where $\bm{\omega}_{i}=[\omega_1^i,\ldots,\omega_{N_i}^i]^\top$, $\lambda_i(t_n^i,\omega_n^i)=\overline{\lambda}_i\text{PG}(\omega_n^i\mid1,0)$ and we write the point process likelihood and state process likelihood as two independent components. 

\subsection{Gibbs Sampler}
Combining the priors in \cref{eq6} and the augmented likelihood in \cref{eq9}, we obtain the augmented joint distribution $p(D,\{\bm{\omega}_i\},\{\Pi_i\},\{\bm{\Phi}_i\},\{\overline{\lambda}_i\},\{\mathbf{w}_{i}^k\})$. Notice that if we marginalize out the latent variables, the resulting marginal will be the same as the original distribution. Based on the augmented joint distribution, we can derive the conditional densities of latent variables and parameters in closed form, which is provided in the appendix. By sampling from these conditional densities iteratively, we construct an analytical Gibbs sampler. 

Although the Gibbs sampler is efficient to some extent due to the closed-form expressions, it is still not efficient enough for practical usage. The bottleneck is the sampling of latent Poisson processes, which is performed by the time-consuming thinning algorithm~\citep{ogata1998space}. To further improve efficiency, a mean-field variational inference is derived to provide an approximated posterior. 

\subsection{Mean-Field Variational Inference}

As mentioned earlier, since the Dirichlet prior is conjugate to the state process likelihood, the state process posterior $p(\bm{\Phi}_i\mid D)$ can be solved analytically 
\begin{equation}
p(\bm{\Phi}_i\mid D)=\prod_{k=1}^K\text{Dir}(\bm{\phi}_k^i\mid \mathbf{s}^i_k+\bm{\alpha}),
\label{eq13}
\end{equation}
where $\mathbf{s}_k^i=[s^i_{k,1},\ldots,s^i_{k,K}]$ is the count of transitions from $k$ to $k'$ on $i$-th dimensional point process. 

For the mean-field, we only need to approximate the point process posterior $p(\bm{\omega}_i,\Pi_i,\overline{\lambda}_i,\mathbf{w}_{i}\mid D)$. A variational distribution that is assumed to factorize over some partition of latent variables is optimized to approximate the real posterior. For the current problem, we assume the variational distribution of $i$-th dimensional point process factorizes as $q(\bm{\omega}_i,\Pi_i,\overline{\lambda}_i,\mathbf{w}_{i})=q_1(\bm{\omega}_i,\Pi_i)q_2(\overline{\lambda}_i,\mathbf{w}_{i})$. By using the calculus of variations, it can be shown that the optimal distribution for each factor~\citep{bishop2006pattern}, in terms of minimizing the Kullback-Leibler (KL) divergence, can be expressed as 
\begin{equation}
\label{eq11}
\begin{aligned}
q_1(\bm{\omega}_i,\Pi_i)&\propto\exp(\mathbb{E}_{q_2}[\log p(D,\bm{\omega}_i,\Pi_i,\overline{\lambda}_i,\mathbf{w}_{i})]),\\
q_2(\overline{\lambda}_i,\mathbf{w}_{i})&\propto\exp(\mathbb{E}_{q_1}[\log p(D,\bm{\omega}_i,\Pi_i,\overline{\lambda}_i,\mathbf{w}_{i})]).
\end{aligned}
\end{equation}
Substituting the augmented density $p(D,\bm{\omega}_i,\Pi_i,\overline{\lambda}_i,\mathbf{w}_i)$ derived from \cref{eq6,eq9} into \cref{eq11}, we obtain the optimal distribution for each factor
\begin{subequations}
\label{eq12}
\begin{gather}
\label{eq12a}
q_1(\bm{\omega}_{i})=\prod_{n=1}^{N_i}\text{PG}(\omega_n^i\mid 1,\widetilde{h}_i(t_n^i,z(t_n^i))),\\
\label{eq12b}
\Lambda_i^1(t,\omega)=\overline{\lambda}_i^1\sigma(-\widetilde{h}_i(t,z(t)))\text{PG}(\omega\mid 1,\widetilde{h}_i(t,z(t)))\exp{(\frac{1}{2}(\widetilde{h}_i(t,z(t))-\overline{h}_i(t,z(t))))},\\
\label{eq12d}
q_2(\overline{\lambda}_i)=\text{Gamma}(\overline{\lambda}_i\mid N_i+\widetilde{R}_i,T),\\
\label{eq12e}
q_2(\mathbf{w}_{i})=\prod_{k=1}^K\mathcal{N}(\mathbf{w}_i^k\mid \widetilde{\mathbf{m}}_i^k,\widetilde{\bm{\Sigma}}_i^k).
\end{gather}
\end{subequations}

\Cref{eq12a} is the optimal density of P\'{o}lya-Gamma variables where we utilize the tilted P\'{o}lya-Gamma distribution $\text{PG}(\omega\mid b,c)\propto e^{-c^2\omega/2}\text{PG}(\omega\mid b,0)$, $\widetilde{h}_i(t,z(t))=\sqrt{\mathbb{E}[h_i^2(t,z(t))]}$. The subsequent required expectation of $\text{PG}(\omega\mid b,c)$ is $\mathbb{E}[\omega]=\frac{b}{2c}\tanh\frac{c}{2}$. 
\Cref{eq12b} is the intensity of the optimal marked Poisson processes where $\overline{\lambda}_i^1=e^{\mathbb{E}[\log\overline{\lambda}_i]}$ and $\overline{h}_i(t,z(t))=\mathbb{E}[h_i(t,z(t))]$. 
\Cref{eq12d} is the optimal density of intensity upper-bounds where $\widetilde{R}_i=\int_0^T\int_0^\infty\Lambda_i^1(t,\omega)d\omega dt$, which can be solved by Gaussian quadrature \citep{golub1969calculation}. The required expectation in \cref{eq12b} is $\mathbb{E}[\log\overline{\lambda}_i]=\psi(N_i+\widetilde{R}_i)-\log(T)$ where $\psi(\cdot)$ is the digamma function. 

\Cref{eq12e} is the optimal density of activation weights $\mathbf{w}_{i}=\{\mathbf{w}_{i}^k\}_{k=1}^K$ where $\widetilde{\bm{\Sigma}}_i^k=[\int_{t\in k}A_i(t)\mathbf{F}(t)\mathbf{F}^\top(t)dt+\mathbf{K}^{-1}]^{-1}$, $\widetilde{\mathbf{m}}_i^k=\widetilde{\bm{\Sigma}}_i^k\int_{t\in k}B_i(t)\mathbf{F}(t)dt$ with $A_i(t)=\sum_{n=1}^{N_i}\mathbb{E}[\omega_n^i]\delta(t-t_n^i)+\int_0^\infty\omega\Lambda_i^1(t,\omega)d\omega$ and $B_i(t)=\frac{1}{2}\sum_{n=1}^{N_i}\delta(t-t_n^i)-\frac{1}{2}\int_0^\infty\Lambda_i^1(t,\omega)d\omega$. The integrals for $\widetilde{\bm{\Sigma}}_i^k$ and $\widetilde{\mathbf{m}}_i^k$ are w.r.t. $t\in k$ which means the time intervals with state $k$. All intractable integrals can be solved by Gaussian quadrature. The required expectations in \cref{eq12a,eq12b} are $\mathbb{E}[h_i(t,z(t)=k)]=\mathbf{F}^\top(t)\widetilde{\mathbf{m}}_i^k$ and $\mathbb{E}[h_i^2(t,z(t)=k)]=(\mathbf{F}^\top(t)\widetilde{\mathbf{m}}_i^k)^2+\mathbf{F}^\top(t)\widetilde{\bm{\Sigma}}_i^k\mathbf{F}(t)$. 


Computing the posterior of $\bm{\Phi}_i$ by \cref{eq13} directly and updating the posterior of $\bm{\omega}_i,\Pi_i,\overline{\lambda}_i,\mathbf{w}_i$ iteratively by \cref{eq12}, we obtain a mean-field variational inference algorithm. The mean-field algorithm is faster than the Gibbs sampler because we compute the expectation rather than sampling. 

\paragraph{Complexity}
We define the number of points on all dimensions to be $N$ 
and the number of Gaussian quadrature nodes to be $R_{gq}$. The computational complexity of the mean-field algorithm is $\mathcal{O}(L((N+MR_{gq})(MB+1)^2+KM(MB+1)^3))$ where $L$ is the number of iterations. The complexity of Gibbs sampler is provided in the appendix. 

\paragraph{Hyperparameters}
The hyperparameters of the mean-field algorithm comprise the prior hyperparameters $\bm{\alpha}$ and $\mathbf{K}$, the support of influence function $T_f$, the number and parameters of basis functions and the number of Gaussian quadrature nodes. In experiments, $\bm{\alpha}$ is set to $\mathbf{1}$ to represent a uniform Dirichlet prior; $\mathbf{K}$ is simplified as a diagonal matrix $\sigma^2\mathbf{I}$ where $\mathbf{I}$ is the identity matrix; $\sigma^2$, $T_f$, the number and parameters of basis functions are chosen by cross validation. The number of quadrature nodes is a trade-off between accuracy and efficiency, and is set to a suitable value by trial and error.

\section{Experiments}

In this section, we first conduct experiments to compare Gibbs with mean-field, and then compare our proposed FS-Hawkes with state-of-the-art Hawkes process models. 

\subsection{Comparison between Gibbs and Mean-Field}
\label{sec4.1}

\begin{wraptable}{r}{0.41\columnwidth}\vspace{-.45cm}
\caption{The statistics of simulated (SIM), SCE and INTC datasets. We show the number of dimensions, states and events (training/test) in each dataset.}
\label{tab3}
\begin{center}
\scalebox{0.7}{
\begin{sc}
\begin{tabular}{lcccr}
\toprule
Dataset & Dimension & States & \# of Events\\
\midrule
SIM & 2 & 2 & 5590 / 5749\\
SCE & 1 & 2 & 2100 / 468\\
INTC & 2 & 3 & 4938 / 5344\\
\bottomrule
\end{tabular}
\end{sc}}
\end{center}
\vskip -0.2in
\end{wraptable}

In this section, we compare the accuracy and efficiency between Gibbs sampler and mean-field variational inference. We simulate a 2-state 2-dimensional \textit{self-exciting} and \textit{mutual-inhibitive} FS-Hawkes on $[0,T=2,000]$ with 2 basis functions $\widetilde{f}_{1,2}$ being scaled and shifted Beta densities on $[0,T_f=6]$. 
We use the thinning algorithm~\citep{ogata1998space} to simulate two sets of data as the training and test datasets, which contain 5590 and 5749 events respectively. Details about the simulation procedure are provided in the appendix. The simulated training dataset on $[0,200]$ is shown in \cref{fig1} where the state process switches between two states and the event dynamics are temporally heterogeneous according to the state. The statistics of the simulated data are summarized in \cref{tab3}. In the following, we use the superscript in brackets to indicate states and the subscript to indicate dimensions or basis functions.

\paragraph{Results}
We use the proposed Gibbs sampler and mean-field variational inference to perform inference on the training data. Details about training are provided in the appendix. The estimated influence functions $\hat{f}_{11}$ at two states are shown in \cref{fig2} (other estimated influence functions are shown in the appendix). It is easy to see that both algorithms successfully recover the ground truth at two states. The posterior variance of the mean-field algorithm is lower than that of the Gibbs sampler, which is a well-known result in \cite{blei2017variational}. 

\begin{figure}[t]
\begin{center}
\hspace{0.02in}
\begin{minipage}{0.38\columnwidth}
\includegraphics[width=\columnwidth]{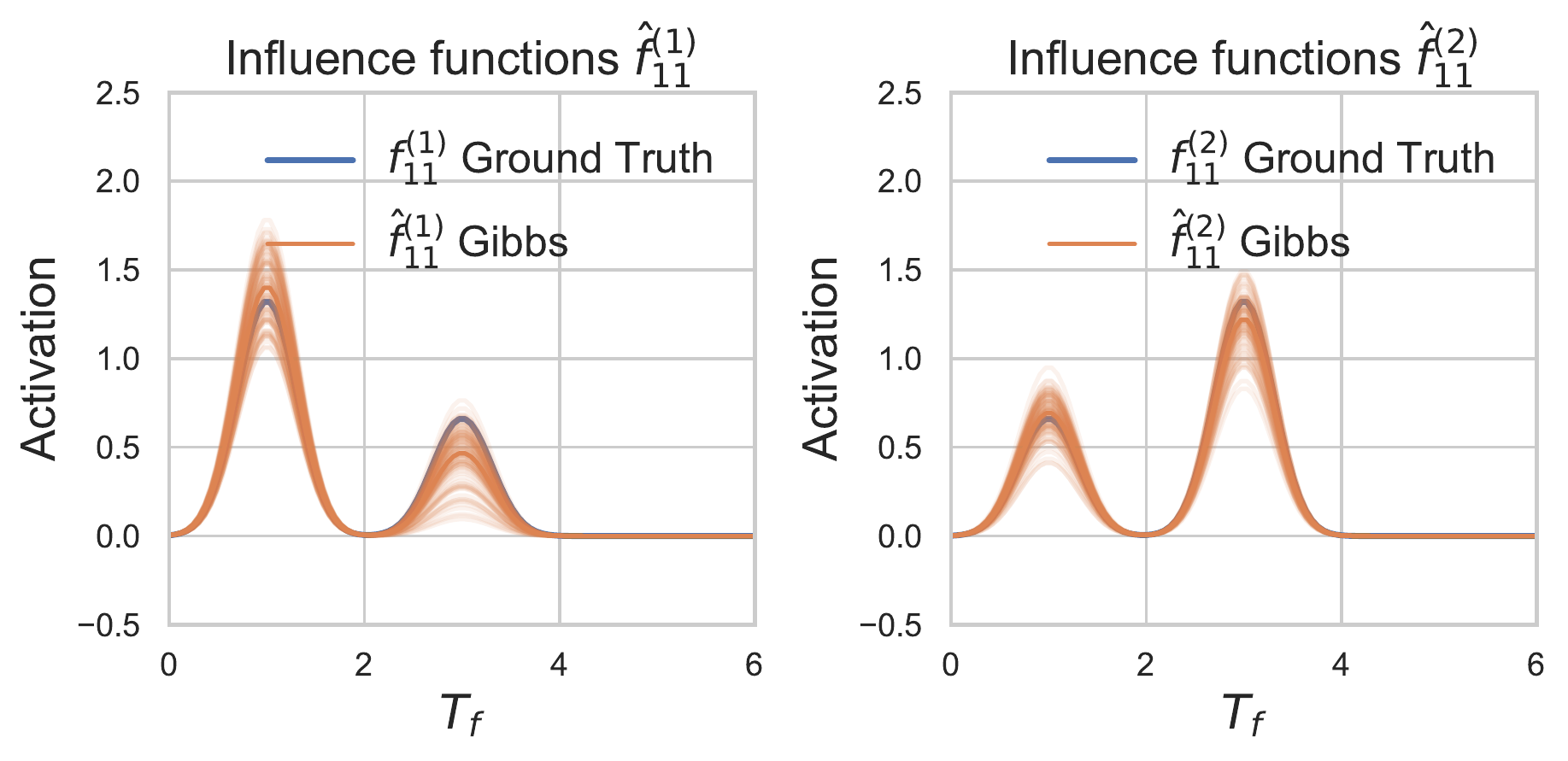}
\subcaption{Gibbs}\label{fig2a}
\end{minipage}
\hspace{0.03in}
\begin{minipage}{0.38\columnwidth}
\includegraphics[width=\columnwidth]{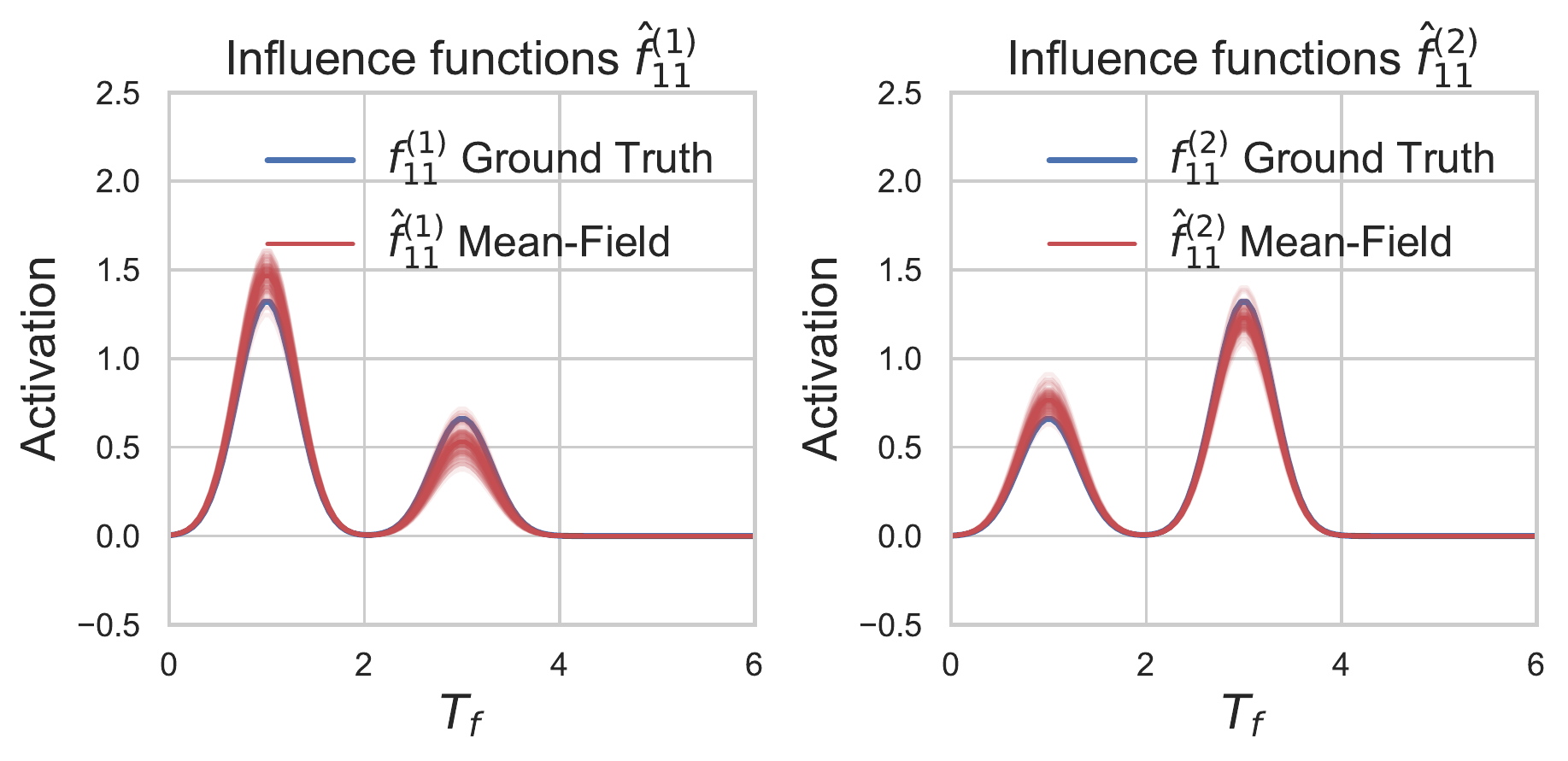}
\subcaption{Mean-Field}\label{fig2b}
\end{minipage}
\hspace{0.01in}
\begin{minipage}{0.2\columnwidth}
\includegraphics[width=\columnwidth]{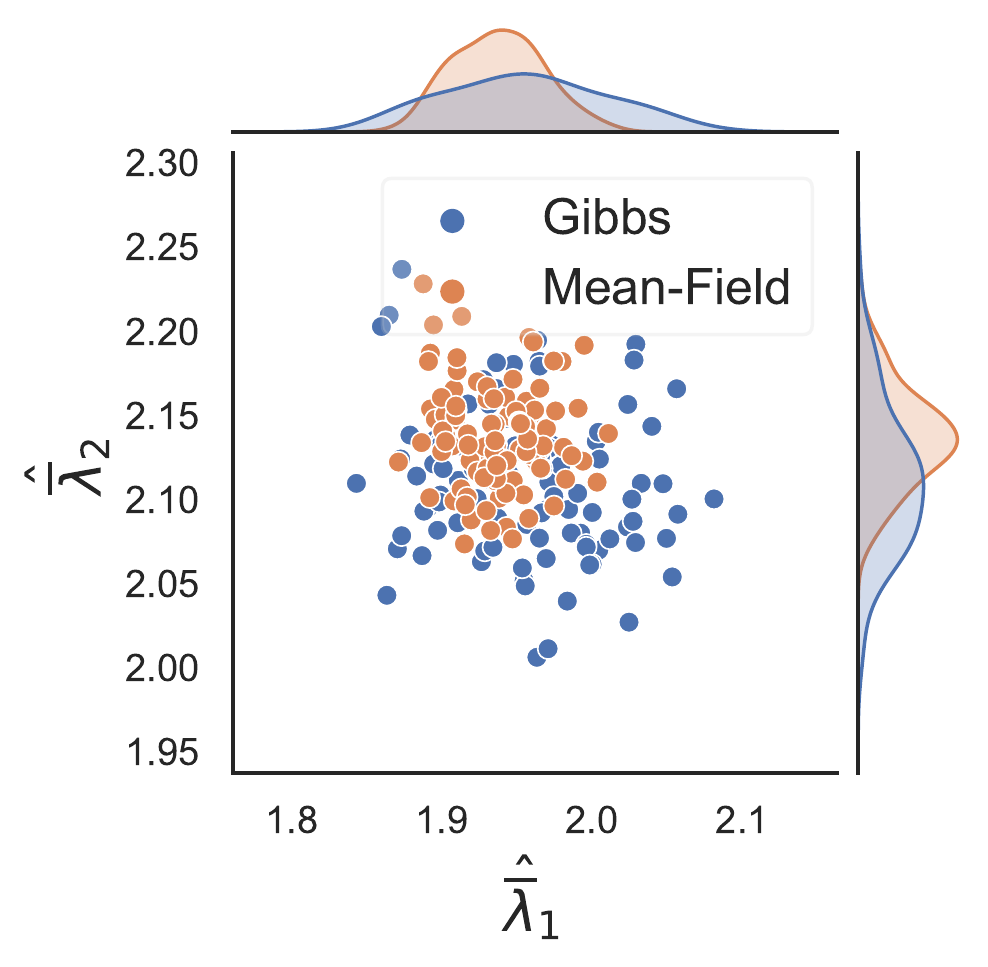}
\subcaption{$\overline{\bm{\lambda}}$}\label{fig2c}
\end{minipage}
\begin{minipage}[b]{0.2\columnwidth}
\includegraphics[width=\columnwidth]{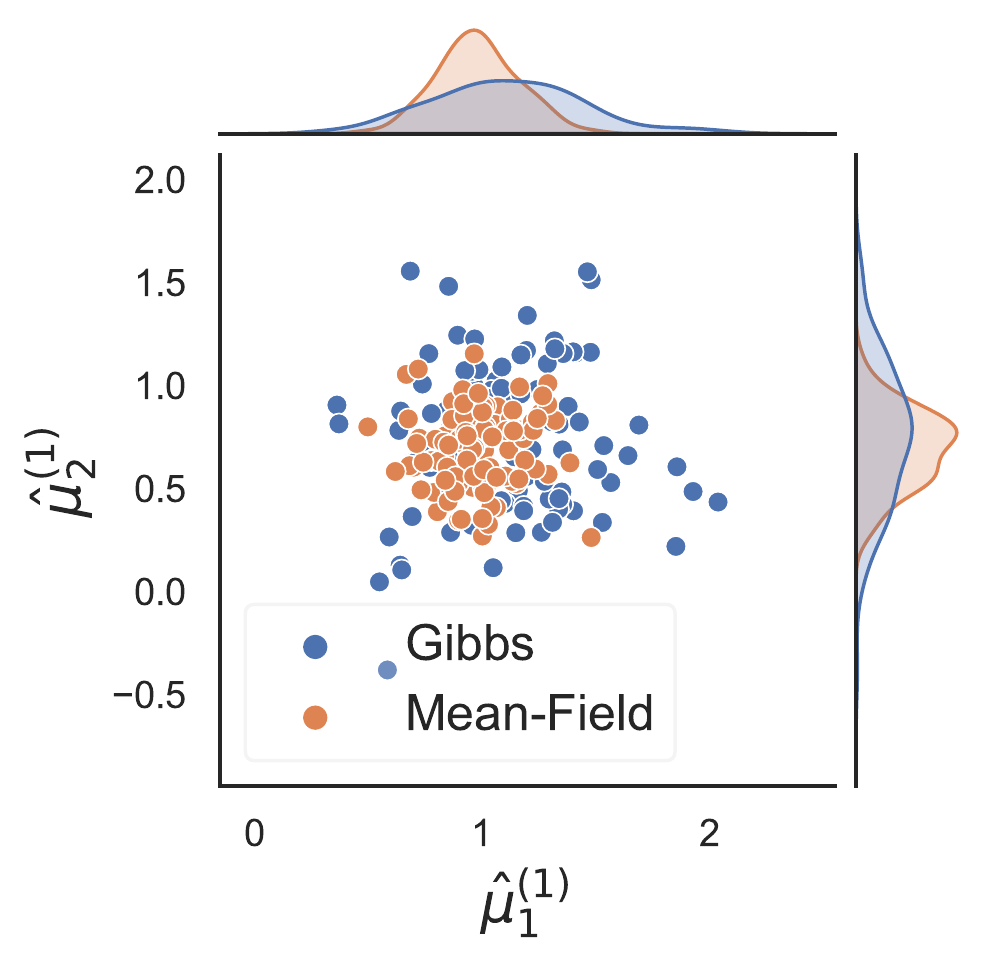}
\subcaption{$\bm{\mu}^{(1)}$}\label{fig2d}
\end{minipage}%
\begin{minipage}[b]{0.2\columnwidth}
\includegraphics[width=\columnwidth]{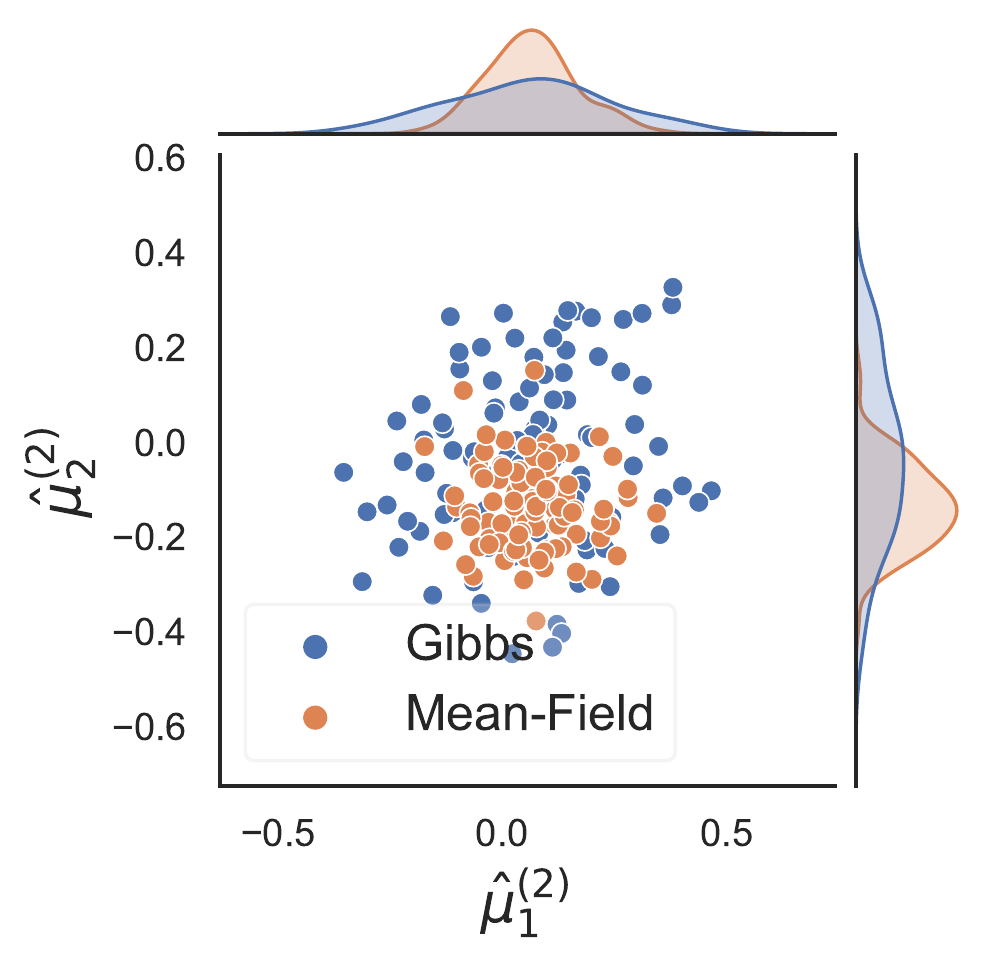}
\subcaption{$\bm{\mu}^{(2)}$}\label{fig2e}
\end{minipage}%
\begin{minipage}[b]{0.2\columnwidth}
\includegraphics[width=\columnwidth]{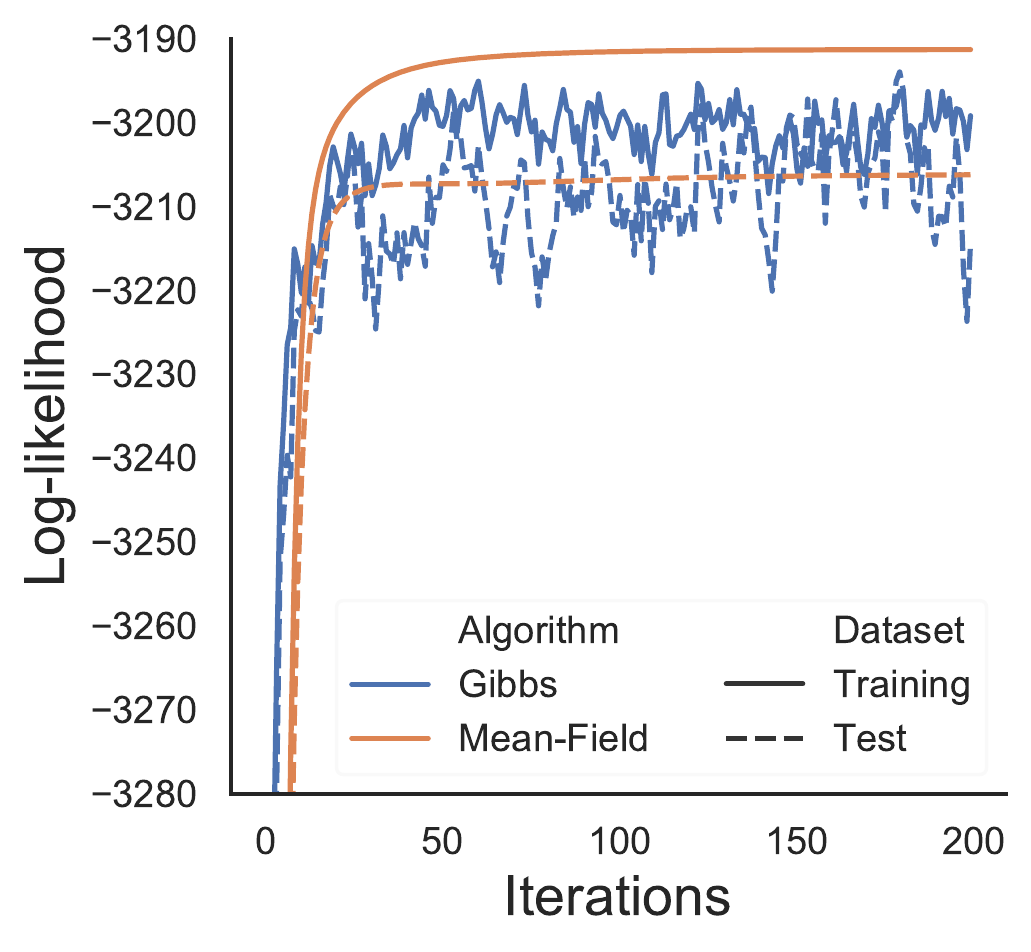}
\subcaption{log-likelihood}\label{fig2f}
\end{minipage}%
\begin{minipage}[b]{0.18\columnwidth}
\includegraphics[width=\columnwidth]{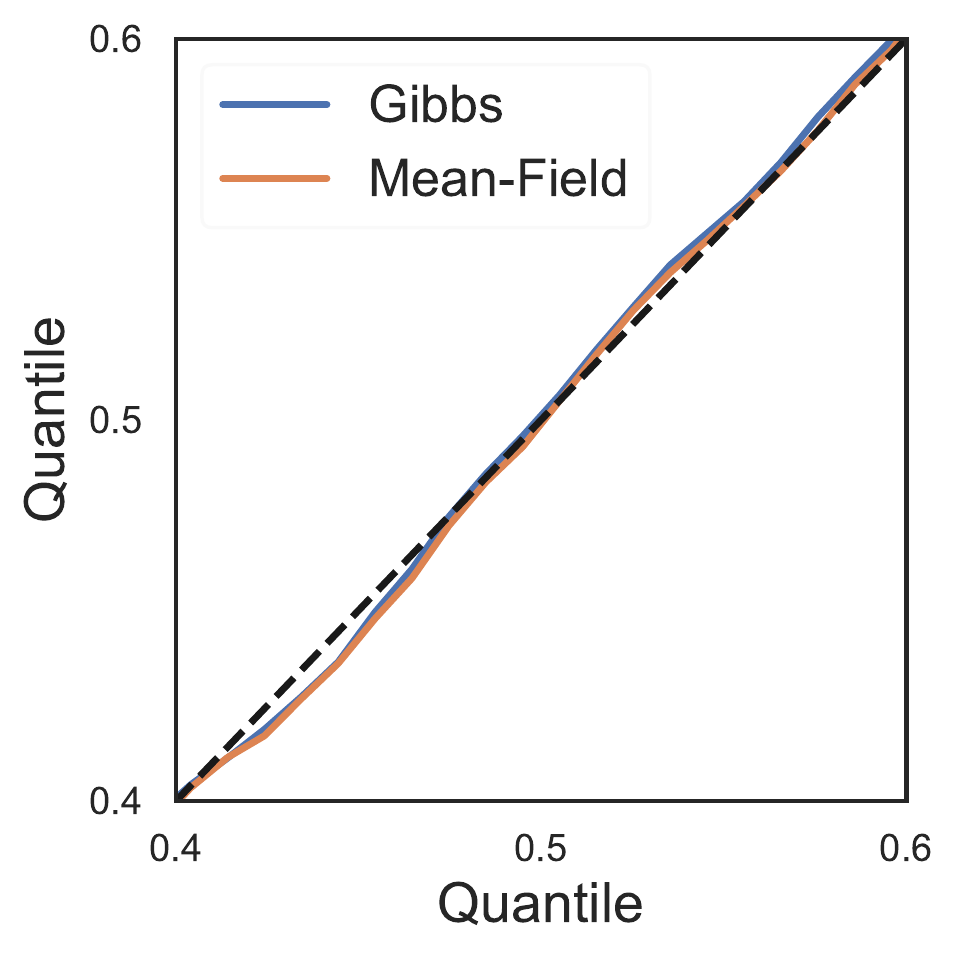}
\subcaption{Q-Q plot}\label{fig2g}
\end{minipage}%
\begin{minipage}[b]{0.18\columnwidth}
\includegraphics[width=\columnwidth]{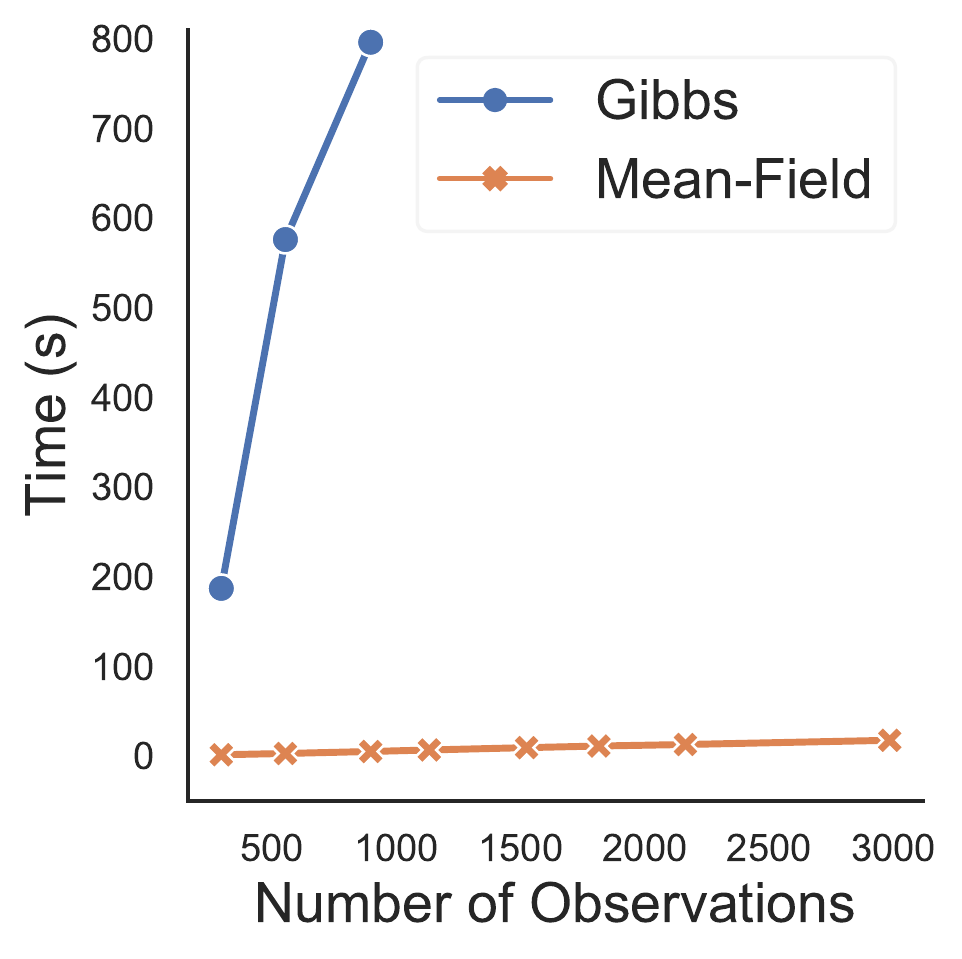}
\subcaption{running time}\label{fig2h}
\end{minipage}
\caption{The $100$ posterior samples of the influence function $\hat{f}_{11}$ at two states from the (a) Gibbs sampler and (b) mean-field variational inference. $\hat{f}_{11}^{(1)}$ has a large bump at the beginning with a small one afterwards while $\hat{f}_{11}^{(2)}$ is the other way around. The $100$ posterior samples of (c) intensity upper-bounds $\overline{\bm{\lambda}}$ and base activations (d) $\bm{\mu}^{(1)}$, (e) $\bm{\mu}^{(2)}$ from the Gibbs sampler and mean-field variational inference. (f) is the training/test log-likelihood curves of both algorithms (for mean-field, it is evaluated by the mean). (g) is the Q-Q plot of both algorithms on the 1-st dimensional test data (zoom in on $[0.4,0.6]$). (h) is the running time of both algorithms w.r.t. the number of observations.}
\label{fig2}
\end{center}
\vskip -0.3in
\end{figure}

\begin{wraptable}{r}{0.5\columnwidth}\vspace{-.4cm}
\caption{The estimation statistics of all parameters for the Gibbs and mean-field based on 100 posterior samples. The mean and standard deviation (in brackets) are provided; the mean closer to ground truth is emphasized. The state-transition matrices have the same posterior in both algorithms, so the empirical statistics are almost the same in two cases and we only show the mean.}
\label{tab2}
\begin{center}
\scalebox{0.63}{
\begin{sc}
\begin{tabular}{lcccr}
\toprule
Parameters & Ground Truth & Gibbs & Mean-Field\\
\midrule
$\hat{\overline{\lambda}}_1$ & 2.0 & \textbf{1.955}(0.053) & 1.937(0.029)\\
$\hat{\overline{\lambda}}_2$ & 2.0 & \textbf{2.111}(0.043) & 2.138(0.030)\\
$\hat{\mu}_1^{(1)}$ & 1.0 & 1.109(0.325) & \textbf{0.978}(0.178)\\
$\hat{\mu}_2^{(1)}$ & 1.0 & \textbf{0.752}(0.352) & 0.694(0.186)\\
$\hat{\mu}_1^{(2)}$ & 0.0 & \textbf{0.060}(0.177) & 0.065(0.097)\\
$\hat{\mu}_2^{(2)}$ & 0.0 & \textbf{-0.032}(0.181) & -0.133(0.088)\\
$\bm{\Phi}_1$ & $\begin{bmatrix}  0.99 & 0.01 \\ 0.01 & 0.99 \end{bmatrix}$ & $\begin{bmatrix}  0.98 & 0.019 \\ 0.009 & 0.99 \end{bmatrix}$ & $\begin{bmatrix}  0.98 & 0.019 \\ 0.009 & 0.99 \end{bmatrix}$\\
$\bm{\Phi}_2$ & $\begin{bmatrix}  0.99 & 0.01 \\ 0.01 & 0.99 \end{bmatrix}$ & $\begin{bmatrix}  0.99 & 0.009 \\ 0.009 & 0.99 \end{bmatrix}$ & $\begin{bmatrix}  0.99 & 0.009 \\ 0.009 & 0.99 \end{bmatrix}$\\
\bottomrule
\end{tabular}
\end{sc}}
\end{center}
\vskip -0.4in
\end{wraptable}

The posterior samples of intensity upper-bounds $\overline{\lambda}_1$, $\overline{\lambda}_2$ and base activations $\mu_1^{(1)}$, $\mu_2^{(1)}$, $\mu_1^{(2)}$ and $\mu_2^{(2)}$ are shown in \cref{fig2c,fig2d,fig2e}. Again, both algorithms recover the ground truth and the posterior variance of the mean-field is lower than that of the Gibbs. The statistics of the estimated $\{\hat{\overline{\lambda}}\}$, $\{\hat{\mu}\}$ and $\{\hat{\bm{\Phi}}\}$ are shown in \cref{tab2} where we can see the Gibbs is slightly more accurate than the mean-field but with a larger variance.

We also compare the training/test log-likelihood of Gibbs and mean-field. The training/test log-likelihood curves (for mean-field, it is evaluated by the mean) are shown in \cref{fig2f} where the Gibbs and mean-field converge to a near plateau. Besides, we perform the residual analysis \citep{daley2003introduction} which is used to assess the goodness-of-fit. If a point process with intensity $\lambda(t)$ is rescaled as $\tau_i=\int_0^{t_i}\lambda(u)du$, the rescaled times $\{\tau_i\}$ follow a Poisson process with unit rate, which can be visualized by a quantile-quantile (Q-Q) plot. 
The Q-Q plot of Gibbs and mean-field (evaluated by the mean) on the 1-st dimensional test data is shown in \cref{fig2g}, where both algorithms are close to the diagonal indicating a similar goodness-of-fit. 

For efficiency, we plot the running time of both algorithms w.r.t. the number of observations in \cref{fig2h}. As we expected, the Gibbs is less efficient than the mean-field because the sampling is a time-consuming operation. Besides, the running time w.r.t. the number of dimensions $M$, basis functions $B$ and states $K$ is shown in the appendix indicating the superior efficiency of the mean-field. 

Conclusively, both algorithms provide estimations close to the ground truth; the Gibbs sampler is slightly more accurate than the mean-field algorithm as the former accurately characterizes the posterior while the latter provides an approximated one. However, the mean-field algorithm has its own merit on the inference efficiency. In practice, the Gibbs is recommended if accuracy is the first consideration while the mean-field should be given priority if efficiency is important.

\subsection{Comparison with State of the Arts}
In this section, we compare our proposed FS-Hawkes with cutting-edge multi-dimensional Hawkes process models in recent years. 
The baseline models include \textbf{(1)} the \textit{neural HPs (NE-Hawkes)}~\citep{mei2017neural} which is a deep nonlinear HP model where the nonlinear map is a scaled softplus function and the activation is modeled by an LSTM; \textbf{(2)} the \textit{Transformer HPs (TR-Hawkes)}~\citep{simiao2020transformer} whose framework is similar with NE-Hawkes except that the activation is modeled by a Transformer architecture; 
\textbf{(3)} the \textit{nonlinear HPs (NL-Hawkes)}~\citep{zhou2020efficient} that is a flexible, nonlinear and homogeneous version estimated by an expectation–maximization algorithm; \textbf{(4)} the \textit{state-dependent HPs (SD-Hawkes)}~\citep{morariu2018state} which is a parametric (exponential decay), linear and nonhomogeneous version estimated by maximum likelihood estimation. We conduct experiments on two traditional Hawkes process application domains: \textit{seismology} and \textit{high frequency finance}.

\begin{figure}[t]
\begin{center}
    \begin{minipage}[b]{0.24\textwidth}
    \includegraphics[width=\columnwidth]{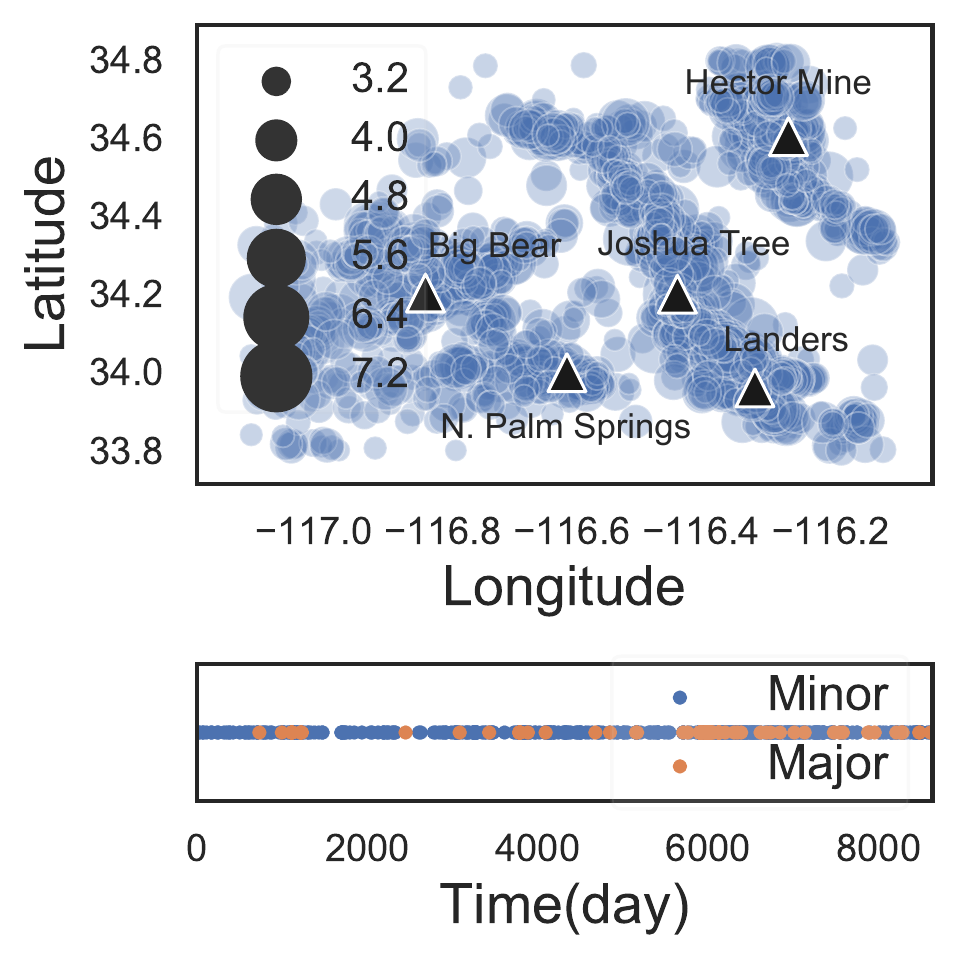}
    \subcaption{}\label{fig4a}
    \end{minipage}%
    \begin{minipage}[b]{0.24\textwidth}
    \includegraphics[width=\columnwidth]{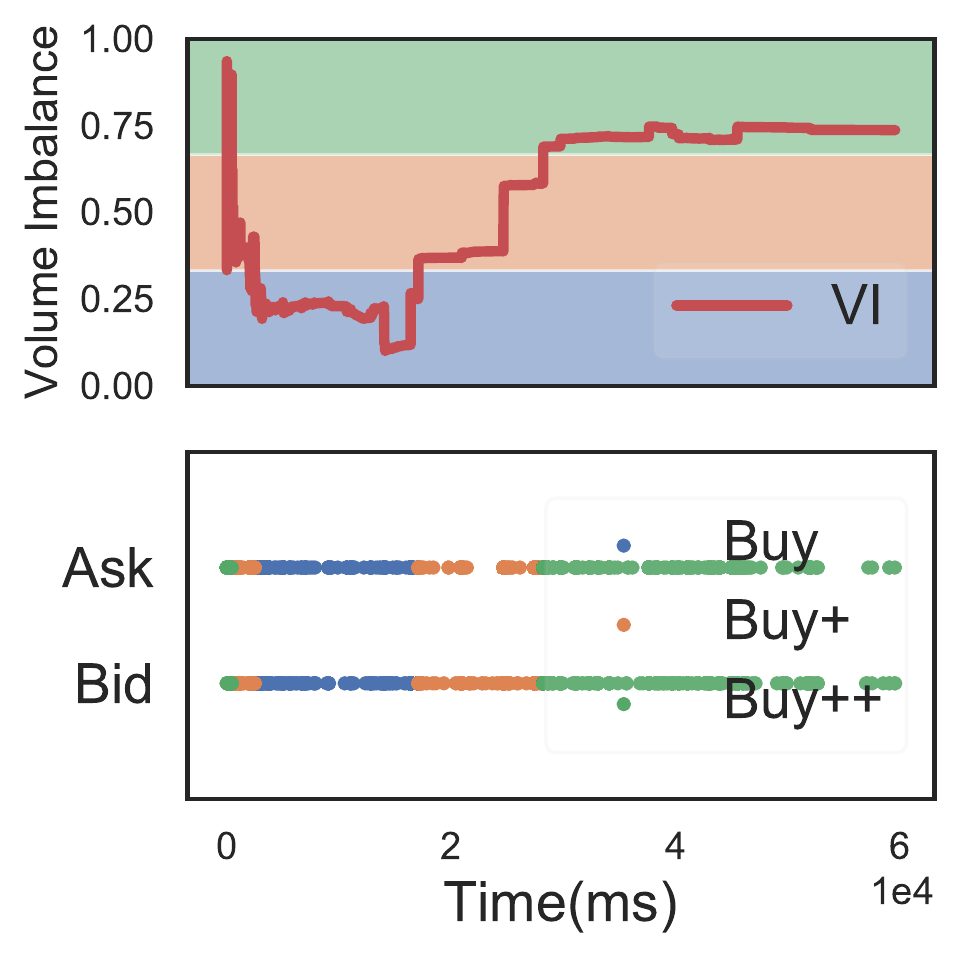}
    \subcaption{}\label{fig4b}
    \end{minipage}%
    \begin{minipage}[b]{0.24\textwidth}
    \includegraphics[width=\columnwidth]{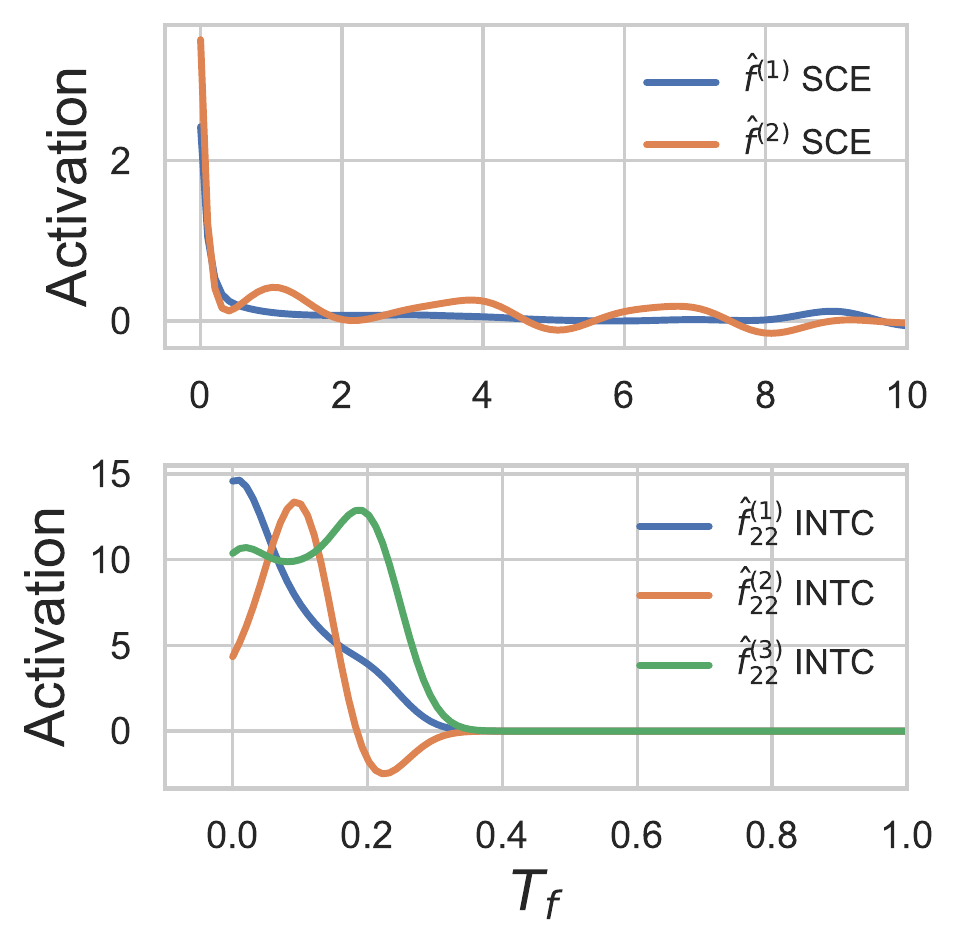}
    \subcaption{}\label{fig4c}
    \end{minipage}%
    \begin{minipage}[b]{0.24\textwidth}
    \includegraphics[width=\columnwidth]{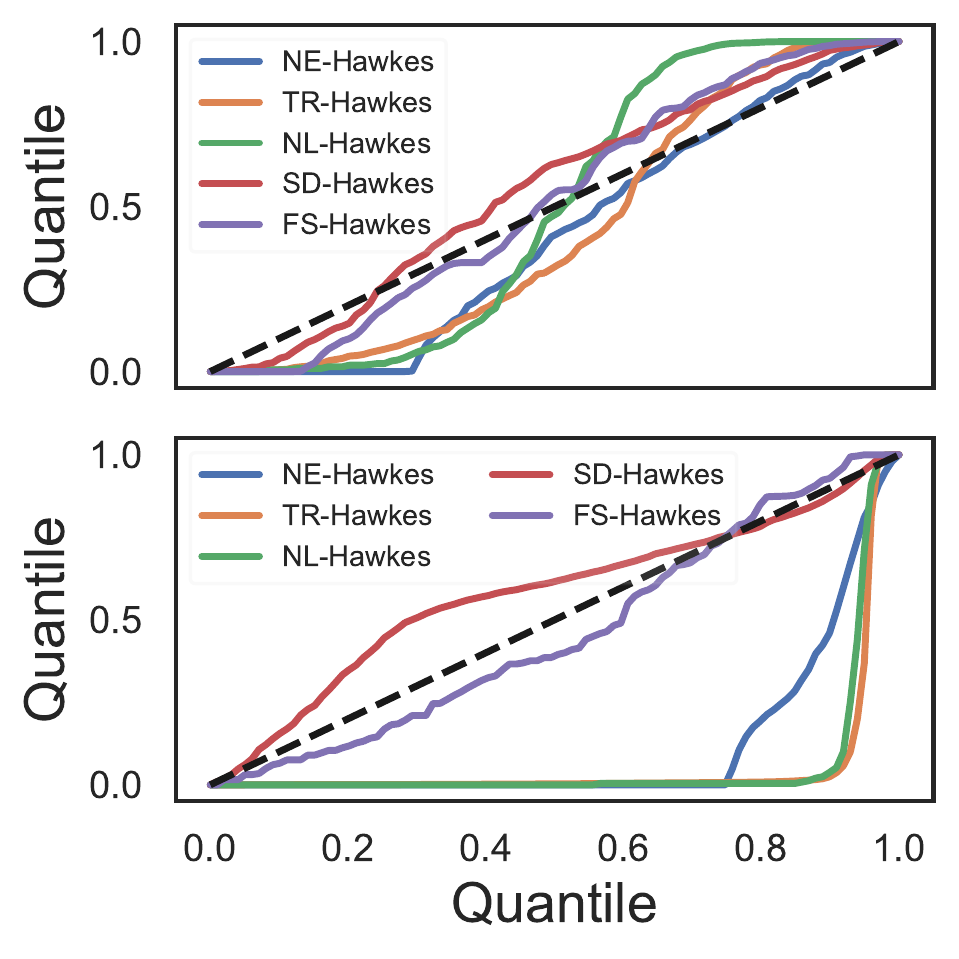}
    \subcaption{}\label{fig4d}
    \end{minipage}
\caption{(a) Locations/magnitudes of earthquakes in the SCE training data with timestamps on the bottom; the significant ones (magnitude$>6$) are plotted as $\blacktriangle$ with name. (b) The volume imbalance through time and timestamps of ask/bid orders in the INTC training data. (c) The estimated influence functions at different states of SCE (top) and INTC (bottom); other estimated influence functions are shown in the appendix. (d) The Q-Q plots on the test data of SCE (top) and INTC (bottom).} 
\label{fig4}
\end{center}
\vskip -0.2in
\end{figure}

\paragraph{Southern California Earthquakes}
The earthquake data is from the southern California earthquake (SCE) data center\footnote{\url{http://www.data.scec.org/index.html}}. Since the Landers earthquake is one of the most powerful earthquakes in southern California and followed by more than ten large aftershocks, the seismic activity in the nearby region around it~\citep{wang2012markov} is analyzed here. 
We collect timestamps of 2568 earthquakes, which are split into training/test data shown in \cref{tab3}. More details about the data are provided in the appendix. Because the earthquakes with different magnitudes have quite different impacts on aftershocks, we divide them into two states: minor (magnitude$\leq4$) and major (magnitude$>4$). As a result, we obtain a 2-state 1-dimensional point process data in a small geographic area; the statistics of the data are summarized in \cref{tab3}. Our goal is to uncover the temporal influence pattern but it can be easily extended to the geographic influence. The locations, magnitudes and timestamps of the earthquakes are plotted in \cref{fig4a}. It is clear that clustering effects exist around those significant earthquakes. 

\paragraph{Level-I LOB of Intel}
In high frequency trading markets, sequential ask and bid orders are separated only by a few microseconds. This raises a requirement to statistically model the price formation at short timescales. An important goal is to model the limit order book (LOB) which is the record of currently unfilled orders. The point process is one of the main approaches to model LOB. The endogenous state of LOB, e.g., the volume imbalance, has a vital impact on the arrival rate of orders because it implies the trend of price change~\citep{morariu2018state}. We collect the level-I LOB data on the stock of Intel (INTC) traded on NASDAQ in one and a half minutes on 2012-06-21\footnote{\url{https://lobsterdata.com}}. We use the order type: ask or bid as the event dimension, the volume imbalance as the state, and split them into training/test data shown in \cref{tab3}. The definition of volume imbalance~\citep{cartea2018enhancing} is $\rho(t)=\frac{V_b(t)-V_a(t)}{V_b(t)+V_a(t)}\in[-1,1]$,
where $V_b(t)$ and $V_a(t)$ are the volumes at time $t$ of limited orders at the best bid and ask price with $\rho\to1\ (-1)$ indicating the buying (selling) pressure respectively. We uniformly divide the interval of $[-1,1]$ into 6 bins representing sell++, sell+, sell, buy, buy+ and buy++ states. As the LOB data is just in one day, there only exists the latter three states in the observation. As a result, we obtain a 3-state 2-dimensional point process data with its statistics summarized in \cref{tab3}. The volume imbalance and timestamps of INTC limited orders are plotted in \cref{fig4b}. 


\paragraph{Results}
We evaluate the models by the log-likelihood and residual analysis on the test data. For the time-invariant baseline models, the training/test data are considered as single-state point processes. Considering the efficiency, the inference of FS-Hawkes is performed by the mean-field algorithm. More experimental details about training (e.g., hyperparameters) are given in the appendix. 

The estimated state-transition matrices of both datasets are provided in \cref{tab4}. For SCE, the transition probability of a major or minor earthquake to a minor one is much larger than that to a major one, which is consistent with common sense that the minor earthquakes are more common than the major ones. For INTC, the probability is mainly concentrated on the transition between same states because the volume imbalance changes gradually through time. 

\begin{wraptable}{r}{0.38\columnwidth}\vspace{-.45cm}
\caption{The estimated state-transition matrices (top) and the test log-likelihood per event of all models (bottom) on SCE and INTC.}
\label{tab4}
\begin{center}
\scalebox{0.67}{
\begin{sc}
    \begin{tabular}{lcc}
    \toprule
        & SCE & INTC\\
    \midrule
    $\bm{\Phi}_1$ & $\begin{bmatrix}  0.899 & 0.101 \\ 0.764 & 0.236 \end{bmatrix}$ & $\begin{bmatrix}  0.997 & 0.003 & 0 \\ 0 & 0.997 & 0.003 \\ 0 & 0.001 & 0.999 \end{bmatrix}$\\
    $\bm{\Phi}_2$ & -- & $\begin{bmatrix}  0.998 & 0.002 & 0 \\ 0.004 & 0.996 & 0 \\ 0 & 0.001 & 0.999 \end{bmatrix}$\\
    \bottomrule
    \end{tabular}
    \end{sc}}

    \scalebox{0.8}{
    \begin{sc}
    \begin{tabular}{lcc}
    \toprule
    Log-Likelihood & SCE & INTC\\
    \midrule
    NE-Hawkes & $-2.764$ & $0.676$ \\
    TR-Hawkes & $-3.446$ & $-2.796$ \\
    NL-Hawkes & $-4.892$ & $-3.647$ \\
    SD-Hawkes & $-2.662$ & $1.227$ \\
    FS-Hawkes & \textbf{-2.602} & \textbf{1.391} \\
    \bottomrule
    \end{tabular}
\end{sc}}
\end{center}
\vskip -0.3in
\end{wraptable}

We provide the log-likelihood results on the test data of each dataset in \cref{tab4}. As we expected, the performance of FS-Hawkes is superior to that of baseline models because it can characterize the time-varying (state-dependent) flexible influence functions which are visualized in \cref{fig4c}. We can see that the estimated influence functions are flexible and vary a lot at different states. Also, we perform the residual analysis to assess the goodness-of-fit of different models. The Q-Q plots of all models on the test data of both datasets are shown in \cref{fig4d}. The FS-Hawkes achieves a better goodness-of-fit by approaching the diagonal more closely. 

Conclusively, our FS-Hawkes is superior to baseline models on both datasets where a common feature is that the system state (magnitude of earthquake or volume imbalance of LOB) plays an essential role in driving event dynamics. This serves as a source of competitive advantage of FS-Hawkes over other baseline models that are unable to represent a time-varying flexible point process system.


\section{Related Works}
\label{relatedwork}
Regarding the parametric limitation, some work utilized the frequentist nonparametric approaches to estimate the base intensities and influence functions (e.g., \cite{bacry2016first,lewis2011nonparametric,marsan2008extending,zhou2013learning}), while some work used Bayesian nonparametric methods and most are based on Gaussian processes (e.g., \cite{zhang2018efficient,zhou2019efficient,zhou2020auxiliary}). 

For the linear limitation, a classic approach is to map the convolution of the timestamps with a causal influence function to a non-negative conditional intensity by a nonlinear function~\citep{apostolopoulou2019mutually,gerhard2017stability,zhou2020efficient}, so the influence functions can be negative to represent the inhibitive effect. 

For the homogeneous limitation, few works have been done to solve the problem. The works related to our paper include~\cite{wang2012markov,wu2019markov,zhou2020fast} (the exogenous category) where the state process evolves independently of the point processes constituting an open-loop framework, and~\cite{morariu2018state} (the endogenous category) where the state switching depends on the occurrence of events forming a closed-loop interaction between the state process and point processes. 



\section{Conclusion}

In this paper, we propose the FS-Hawkes which is a flexible, nonlinear and nonhomogeneous HPs variant. In the novel model, an endogenous Markov state process is incorporated to interact with the point processes constituting a closed-loop framework. Our goal is to empower the vanilla HPs to deal with the time-varying system with flexible excitation-inhibition-mixture influence functions. To address the non-conjugate problem in inference, two classes of auxiliary latent variables are augmented to derive two efficient Bayesian inference algorithms: Gibbs sampler and mean-field variational inference with closed-form expressions. The experimental comparison with state-of-the-art competitors demonstrates that the fitting performance of FS-Hawkes is superior on datasets in which the system state has a vital impact on event dynamics.

\bibliography{example_paper}
\bibliographystyle{icml2021}

\clearpage
\appendix
\setcounter{secnumdepth}{0}
\section{Appendices}
\addcontentsline{toc}{section}{Appendices}
\renewcommand{\thesubsection}{\Alph{subsection}}
\setcounter{secnumdepth}{3}
\setcounter{equation}{0}
\setcounter{figure}{0}
\subsection{Augmentation of Marked Poisson Processes}
In this section, we provide a proof of \cref{eq8} in the paper. Utilizing \cref{eq7} in the paper and the logistic symmetry property $\sigma(x)=1-\sigma(-x)$, the exponential integral term in \cref{eq5} in the paper can be written as
\begin{equation}
\begin{aligned}
&\exp{\left(-\int_0^T\overline{\lambda}_i\sigma(h_i(t,z(t)))dt\right)}=\exp{\left(-\iint\left(1-e^{g(\omega,-h_i(t,z(t)))}\right)\overline{\lambda}_i\text{PG}(\omega\mid 1,0)d\omega dt\right)}. 
\end{aligned}
\label{app.eq1}
\end{equation}

The Campbell's theorem~\cite{kingman2005p} indicates that
\begin{equation}
\begin{aligned}
&\mathbb{E}_{\Pi_{\hat{\mathcal{Z}}}}\left[\exp{\left(\xi H(\Pi_{\hat{\mathcal{Z}}})\right)}\right]=\exp{\left[\int_{\hat{\mathcal{Z}}}\left(e^{\xi h(\mathbf{z},\bm{\omega})}-1\right)\Lambda(\mathbf{z},\bm{\omega})d\bm{\omega}d\mathbf{z}\right]},
\end{aligned}
\label{app.eq2}
\end{equation}
where $\Pi_{\hat{\mathcal{Z}}}=\{(\mathbf{z}_n,\bm{\omega}_n)\}_{n=1}^N$ is a marked Poisson process on the product space $\hat{\mathcal{Z}}=\mathcal{Z}\times \Omega$ with intensity $\Lambda(\mathbf{z},\bm{\omega})=\Lambda(\mathbf{z})p(\bm{\omega}\mid \mathbf{z})$. $\Lambda(\mathbf{z})$ is the intensity for the unmarked Poisson process $\{\mathbf{z}_n\}_{n=1}^N$ with $\bm{\omega}_n\sim p(\bm{\omega}_n\mid \mathbf{z}_n)$ being an independent mark drawn at each $\mathbf{z}_n$. $h(\mathbf{z},\bm{\omega}):\mathcal{Z}\times \Omega\rightarrow \mathbb{R}$ is a real-valued function and $H(\Pi_{\hat{\mathcal{Z}}})=\sum_{(\mathbf{z},\bm{\omega})\in\Pi_{\hat{\mathcal{Z}}}}h(\mathbf{z},\bm{\omega})$ is the sum of $h(\mathbf{z},\bm{\omega})$ over $\Pi_{\hat{\mathcal{Z}}}$. \Cref{app.eq2} holds for any $\xi\in \mathbb{C}$ if $\Lambda(\mathbf{z},\bm{\omega})<\infty$. \Cref{app.eq2} defines the characteristic functional of a marked Poisson process. Substituting \cref{app.eq2} into \cref{app.eq1}, we obtain \cref{eq8} in the paper. 

\subsection{Gibbs Sampler}

Based on the augmented joint distribution obtained from \cref{eq6,eq9} in the paper, we obtain the conditional densities of latent variables and parameters in closed form, which constitutes an analytical Gibbs sampler. The $i$-th dimensional conditional densities are 
\begin{subequations}
\label{eq10}
\begin{gather}
\label{eq10a}
p(\bm{\omega}_{i}\mid D,\mathbf{w}_{i})=\prod_{n=1}^{N_i}\text{PG}(\omega_n^i\mid 1,h_i(t_n^i,z(t_n^i))),\\
\label{eq10b}
\Lambda_i(t,\omega\mid D,\mathbf{w}_i,\overline{\lambda}_i)=\overline{\lambda}_i\sigma(-h_i(t,z(t)))\text{PG}(\omega\mid 1,h_i(t,z(t))),\\
\label{eq10c}
p(\bm{\Phi}_i\mid D)=\prod_{k=1}^K\text{Dir}(\bm{\phi}_k^i\mid \mathbf{s}^i_k+\bm{\alpha}),\\
\label{eq10d}
p(\overline{\lambda}_i\mid D,\Pi_i)=\text{Gamma}(\overline{\lambda}_i\mid N_i+R_i,T),\\
\label{eq10e}
p(\mathbf{w}_{i}\mid D,\bm{\omega}_{i},\Pi_i)=\prod_{k=1}^K\mathcal{N}(\mathbf{w}_i^k\mid \mathbf{m}_i^k,\bm{\Sigma}_i^k).
\end{gather}
\end{subequations}
Sampling iteratively by \cref{eq10}, we obtain a sequence of samples to characterize the posterior. 

\Cref{eq10a} is the conditional posterior of P\'{o}lya-Gamma variables where we utilize the tilted P\'{o}lya-Gamma distribution $\text{PG}(\omega\mid b,c)\propto e^{-c^2\omega/2}\text{PG}(\omega\mid b,0)$ and $\mathbf{w}_{i}=\{\mathbf{w}_{i}^k\}_{k=1}^K$. An efficient sampling method \cite{polson2013bayesian} can be used to sample from the P\'{o}lya-Gamma density. \Cref{eq10b} is the conditional posterior intensity of the marked Poisson process. To sample from it, we first use the thinning algorithm \cite{ogata1998space} to draw the timestamps $\{t_r^i\}_{r=1}^{R_i}$ with the rate $\overline{\lambda}_i\sigma(-h_i(t,z(t)))$ and then draw the corresponding marks $\{\omega_r^i\}_{r=1}^{R_i}$ from $\text{PG}(\omega\mid 1,h_i(t,z(t)))$. \Cref{eq10c} is the posterior of the state-transition matrix where $\mathbf{s}_k^i=[s^i_{k,1},\ldots,s^i_{k,K}]$ is the counts of transitions from $k$ to $k'\in\{1,\ldots,K\}$ on the $i$-th dimensional point process. \Cref{eq10d} is the conditional posterior of the intensity upper-bound where $R_i=\left|\Pi_i\right|$ is the number of points on $\Pi_i$. \Cref{eq10e} is the conditional posterior of the activation weights. We define $\{t_n^{i,k},\omega_n^{i,k}\}_{n=1}^{N_{i,k}}$ to be the observed timestamps and latent marks on the $i$-th dimension with state $k$ and $\{t_r^{i,k},\omega_r^{i,k}\}_{r=1}^{R_{i,k}}$ to be the ones on $\Pi_i$ with state $k$. The covariance matrix $\bm{\Sigma}_i^k=[\mathbf{F}_i^k{\mathbf{D}_i^k}^{-1}{\mathbf{F}_i^k}^\top+\mathbf{K}^{-1}]^{-1}$ where $\mathbf{D}_i^k$ is a diagonal matrix with its first $N_{i,k}$ entries being $\{1/\omega_n^{i,k}\}_{n=1}^{N_{i,k}}$ and the following $R_{i,k}$ entries being $\{1/\omega_r^{i,k}\}_{r=1}^{R_{i,k}}$, $\mathbf{F}_i^k=[\{\mathbf{F}(t_n^{i,k})\}_{n=1}^{N_{i,k}},\{\mathbf{F}(t_r^{i,k})\}_{r=1}^{R_{i,k}}]$, $\mathbf{K}$ is the prior covariance matrix in \cref{eq6} in the paper. The mean $\mathbf{m}_i^k=\bm{\Sigma}_i^k\mathbf{F}_i^k\mathbf{v}_i^k$ where the first $N_{i,k}$ entries of $\mathbf{v}_i^k$ are $1/2$ and the following $R_{i,k}$ entries are $-1/2$. 

\paragraph{Complexity}
We define the number of points on $\{\Pi_i\}_{i=1}^M$ to be $R$ and the average number of points on the support of $T_f$ on all dimensions to be $N_{T_f}$. The computational complexity of the Gibbs sampler is $\mathcal{O}(L( RN_{T_f}B+C_{\text{PG}}(N+R)+C_{\text{TH}}M+(N+R)(MB+1)^2+KM(MB+1)^3))$ where $L$ is the number of iterations, $C_{\text{PG}}$ and $C_{\text{TH}}$ are the complexities of P\'{o}lya-Gamma sampling and thinning algorithm. The sampling of other variables are ignored as they are fast. Each component in the complexity is due to the computation of $\mathbf{F}(t)$ on the points of $\{\Pi_i\}$, the sampling operation, the matrix multiplication and inversion, respectively. We can see that the mean-field algorithm is faster than the Gibbs sampler because the computation of $\mathbf{F}(t)$ on the latent Poisson processes is taken out of iterations and the time-consuming sampling operations are avoided. 

\paragraph{Hyperparameters}
The choice of hyperparameters for the Gibbs sampler is similar with that of the mean-field algorithm.

\subsection{Experiments}

In this section, we provide more details of the experiments. 

\subsubsection{Comparison between Gibbs and Mean-Field}
For the simulation, the basis functions are $\tilde{f}_{\{1,2\}}=\text{Beta}(\tilde{\alpha}=50,\tilde{\beta}=50)$ scaled to $T_f=6$ and shifted by $-2$ and $0$, respectively. The state-dependent influence functions are designed as $f_{11}^{(1)}=\widetilde{f}_1+0.5\widetilde{f}_2$, $f_{12}^{(1)}=-0.5\widetilde{f}_1-0.25\widetilde{f}_2$, $f_{21}^{(1)}=-0.25\widetilde{f}_1-0.5\widetilde{f}_2$, $f_{22}^{(1)}=0.5\widetilde{f}_1+\widetilde{f}_2$ at the first state and $f_{11}^{(2)}=0.5\widetilde{f}_1+\widetilde{f}_2$, $f_{12}^{(2)}=-0.25\widetilde{f}_1-0.5\widetilde{f}_2$, $f_{21}^{(2)}=-0.5\widetilde{f}_1-0.25\widetilde{f}_2$, $f_{22}^{(2)}=\widetilde{f}_1+0.5\widetilde{f}_2$ at the second state with positive indicating excitation and negative indicating inhibition. The state-dependent base activations are $\mu_1^{(1)}=\mu_2^{(1)}=1$ at the first state and $\mu_1^{(2)}=\mu_2^{(2)}=0$ at the second state. The intensity upper-bounds are $\overline{\lambda}_1=\overline{\lambda}_2=2$. The dimension-dependent state-transition matrices are $\bm{\Phi}_1=\bm{\Phi}_2=[[\phi(1,1)=0.99,\phi(1,2)=0.01],[\phi(2,1)=0.01,\phi(2,2)=0.99]]$ that means it has a high probability to keep the original state. 

\begin{figure}[ht]
\begin{center}
\begin{minipage}{0.8\linewidth}
\includegraphics[width=\linewidth]{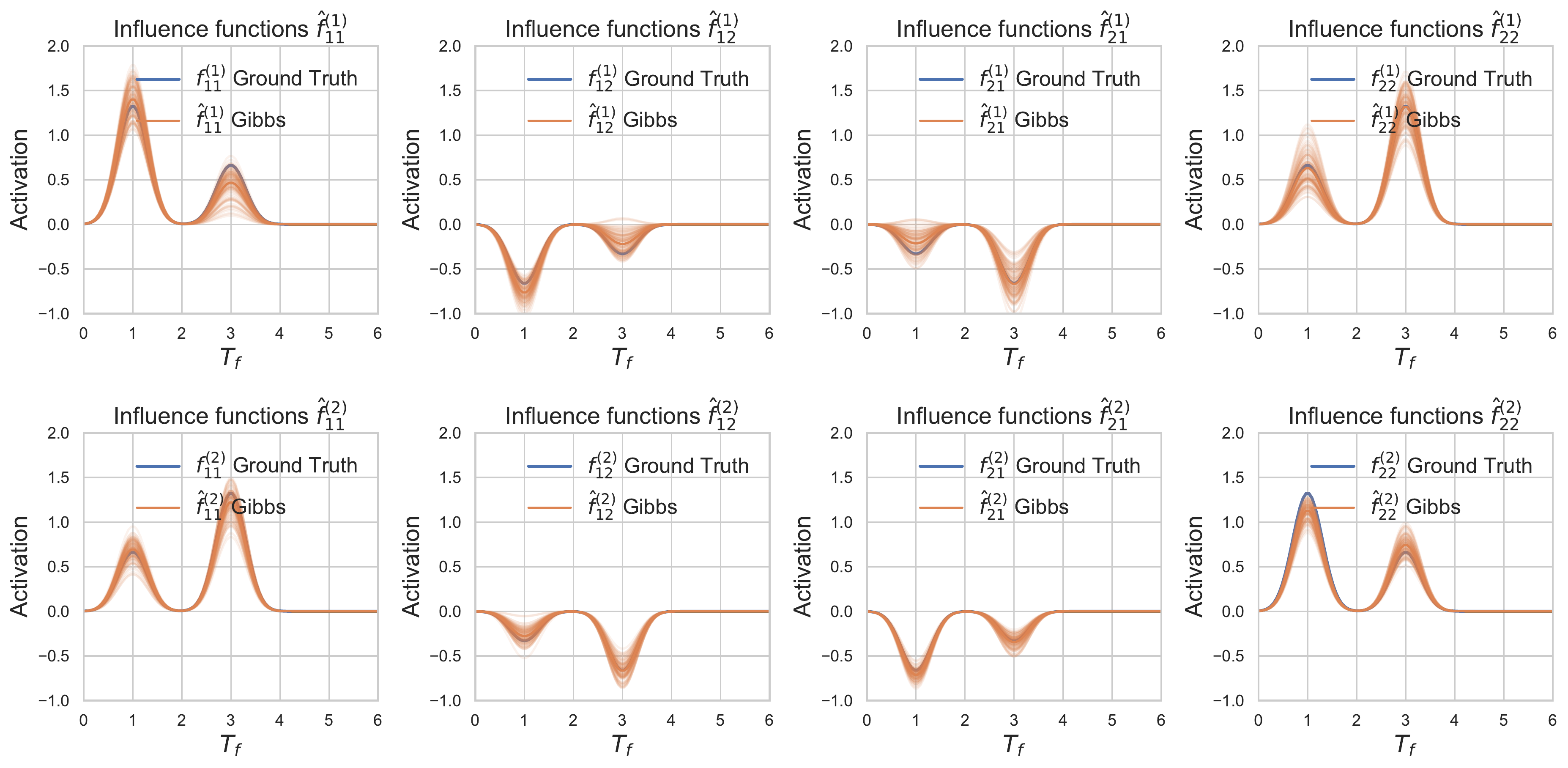}
\subcaption{Gibbs}\label{app.fig2a}
\end{minipage}
\begin{minipage}{0.8\linewidth}
\includegraphics[width=\linewidth]{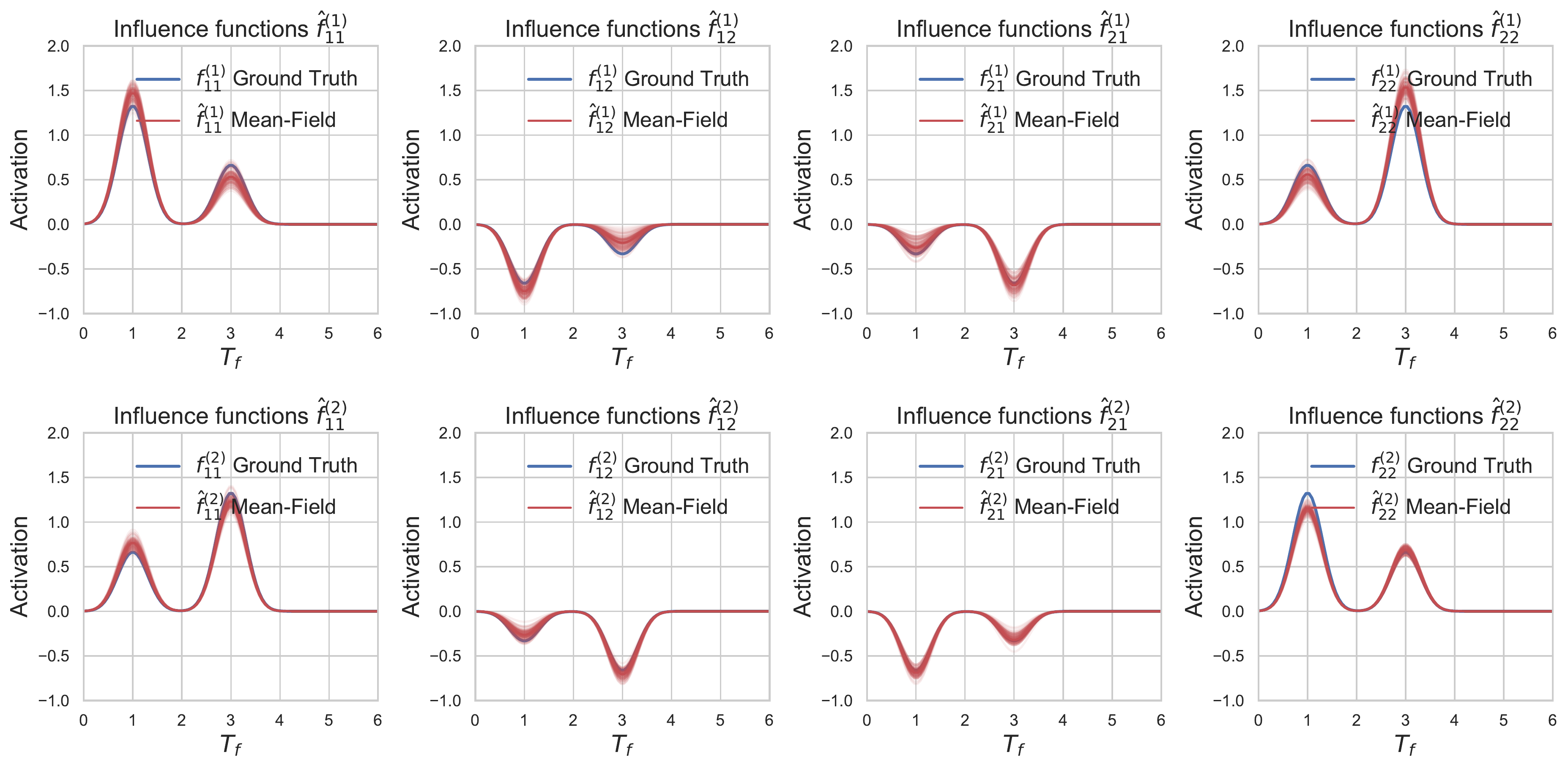}
\subcaption{Mean-Field}\label{app.fig2b}
\end{minipage}
\caption{The $100$ posterior trajectories of all influence functions at two states from (a) Gibbs sampler and (b) mean-field variational inference.}
\label{app.fig2}
\end{center}
\end{figure}

\begin{figure}[ht]
\begin{center}
\begin{minipage}{0.3\linewidth}
\includegraphics[width=\linewidth]{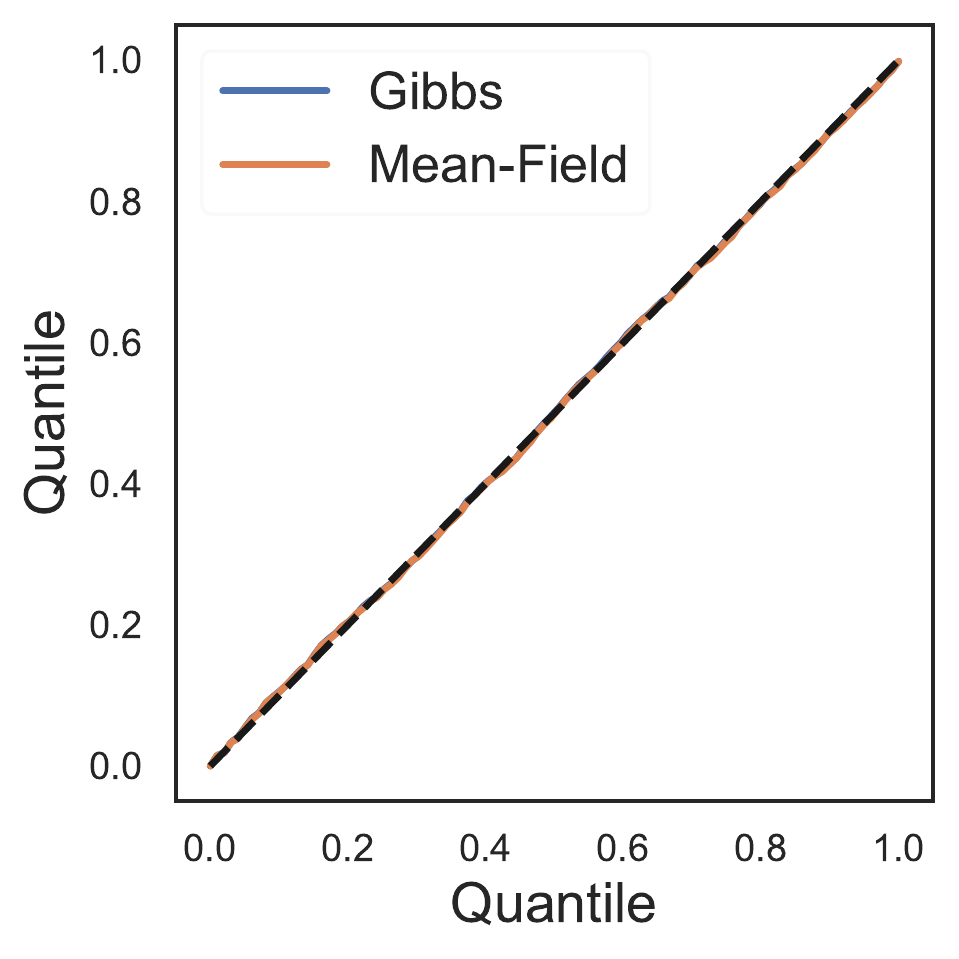}
\subcaption{Dimension 1}\label{app.fig3a}
\end{minipage}%
\begin{minipage}{0.3\linewidth}
\includegraphics[width=\linewidth]{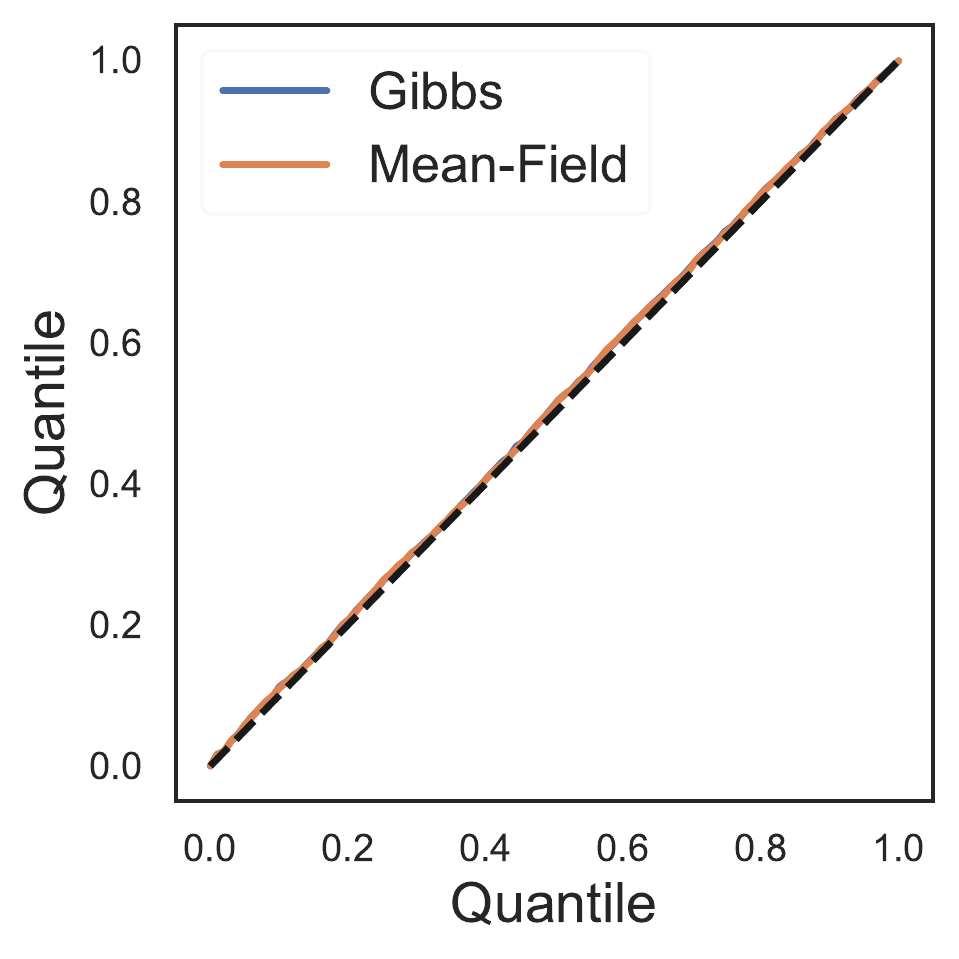}
\subcaption{Dimension 2}\label{app.fig3b}
\end{minipage}
\caption{The Q-Q plots of Gibbs and mean-field on the test data (a) 1-st dimension and (b) 2-nd dimension.}
\label{app.fig3}
\end{center}
\end{figure}

For the training, the basis functions are chosen as the ground truth because it is known. By cross validation, the hyperparameter $\sigma^2$ is chosen to be $1$. The number of grid for integration in the Gibbs sampler is set to 200,000 and the number of quadrature nodes in the mean-field algorithm is set to 100 per state-interval. The number of iterations for both algorithms is set to 200 which is large enough for convergence. All estimated influence functions at two states are shown in \cref{app.fig2}. The Q-Q plots of both algorithms on both dimensions of the test data are shown in \cref{app.fig3}.


\begin{figure}[ht]
\begin{center}
\begin{minipage}[b]{0.3\columnwidth}
\includegraphics[width=\columnwidth]{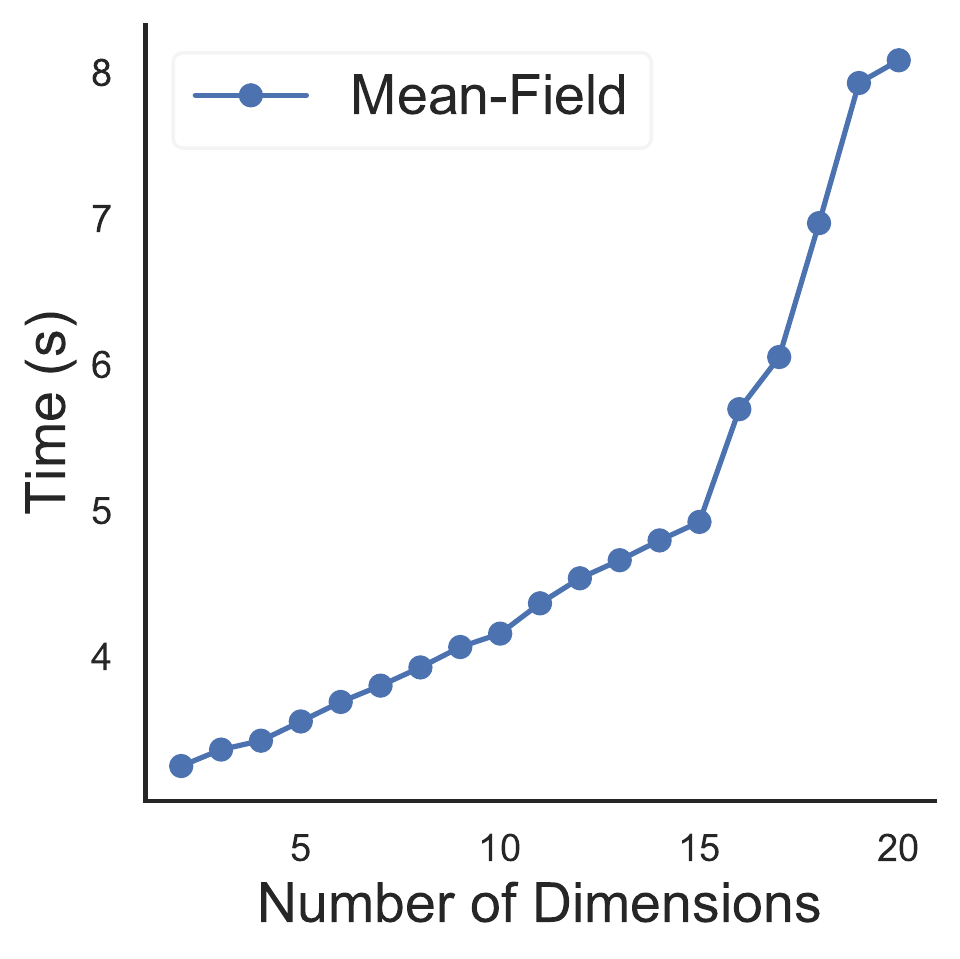}
\end{minipage}%
\begin{minipage}[b]{0.3\columnwidth}
\includegraphics[width=\columnwidth]{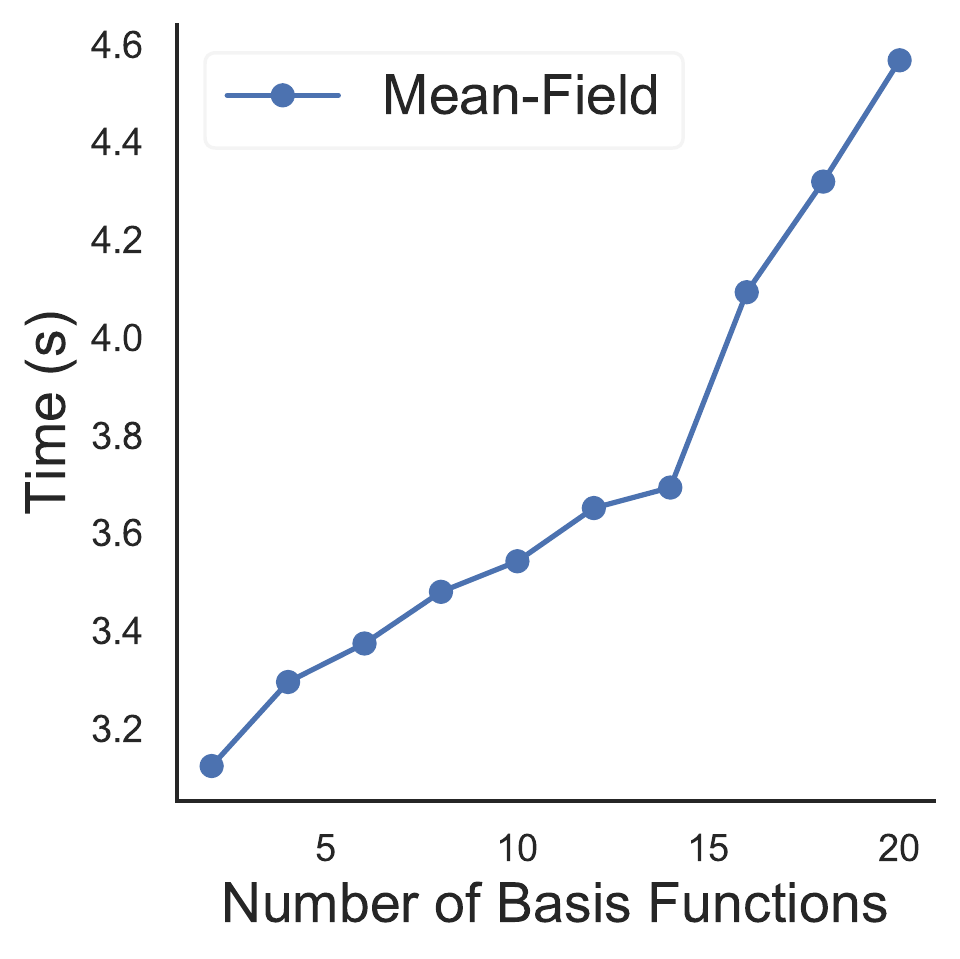}
\end{minipage}%
\begin{minipage}[b]{0.3\columnwidth}
\includegraphics[width=\columnwidth]{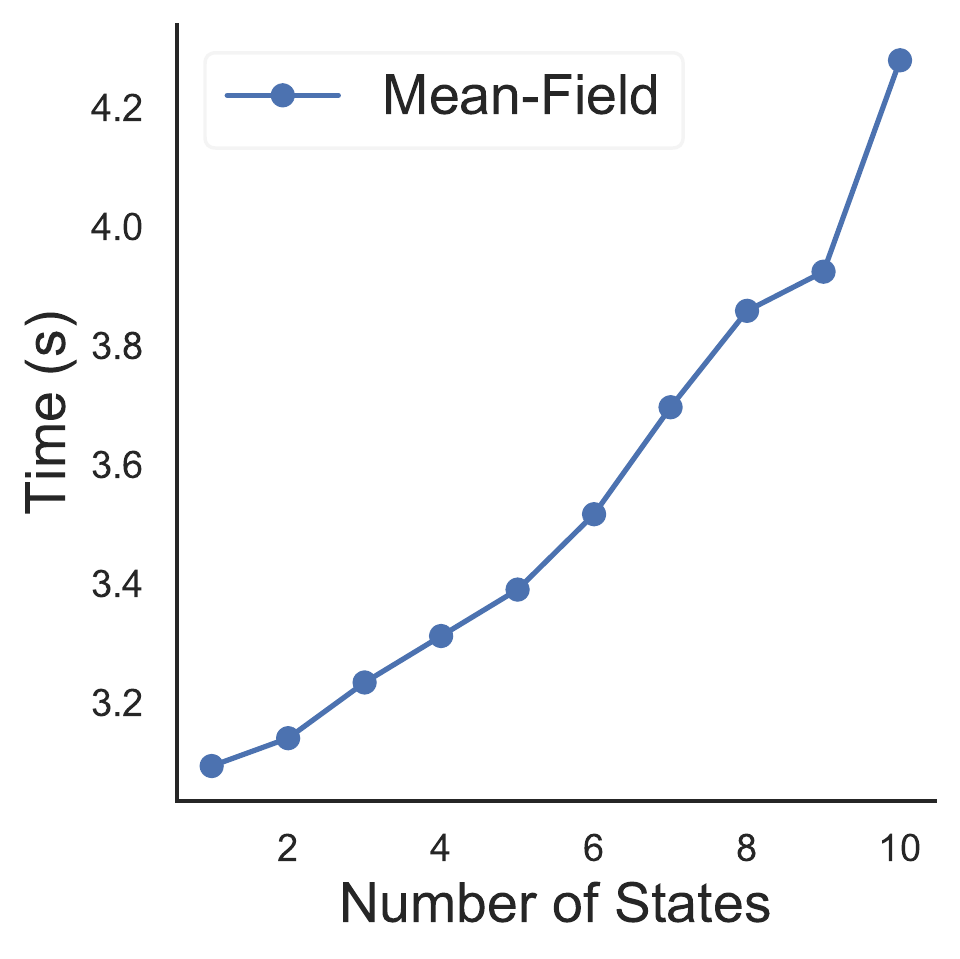}
\end{minipage}
\caption{The running time of mean-filed w.r.t. the number of (a) dimensions ($B=2,K=2$), (b) basis functions ($M=2,K=2$) and (c) states ($M=2,B=2$). The number of observations is 2,000.}
\label{app.fig4}
\end{center}
\vskip -0.2in
\end{figure}

Although our model has many parameters, the inference is efficient. For $\mathbf{w}$ and $\overline{\lambda}$, our posterior inference scheme is conjugate and has closed-form expressions for the iteration; for state-transition matrix $\bm{\Phi}$, the estimation is even more efficient than $\mathbf{w}$ since we have an accurate posterior \cref{eq13} without the need of iteration. We show the running time of our mean-field algorithm w.r.t. the number of dimensions $M$, basis functions $B$ and states $K$ in \cref{app.fig4}. It is clear that the inference is fast for large $M$, $K$ or $B$. Besides, we test the efficiency performance on a dataset with reasonably large $M=20$, $K=10$, $B=6$ and $2000$ observed events; the running time is only $33.24$ seconds showing its superior efficiency. 

\subsubsection{Comparison with State of the Arts}

\begin{figure}[ht]
\begin{center}
\begin{minipage}[b]{0.32\linewidth}
\includegraphics[width=\linewidth]{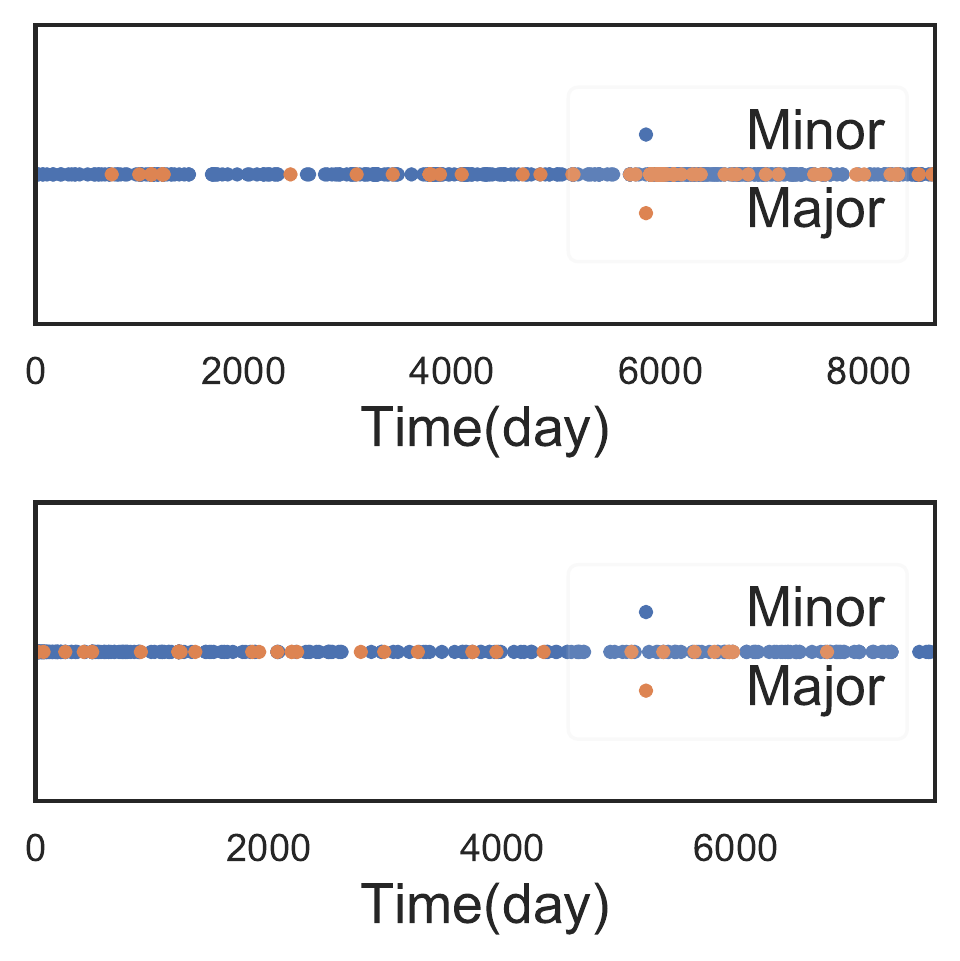}
\subcaption{SCE}\label{app.fig1a}
\end{minipage}%
\begin{minipage}[b]{0.65\linewidth}
\includegraphics[width=\linewidth]{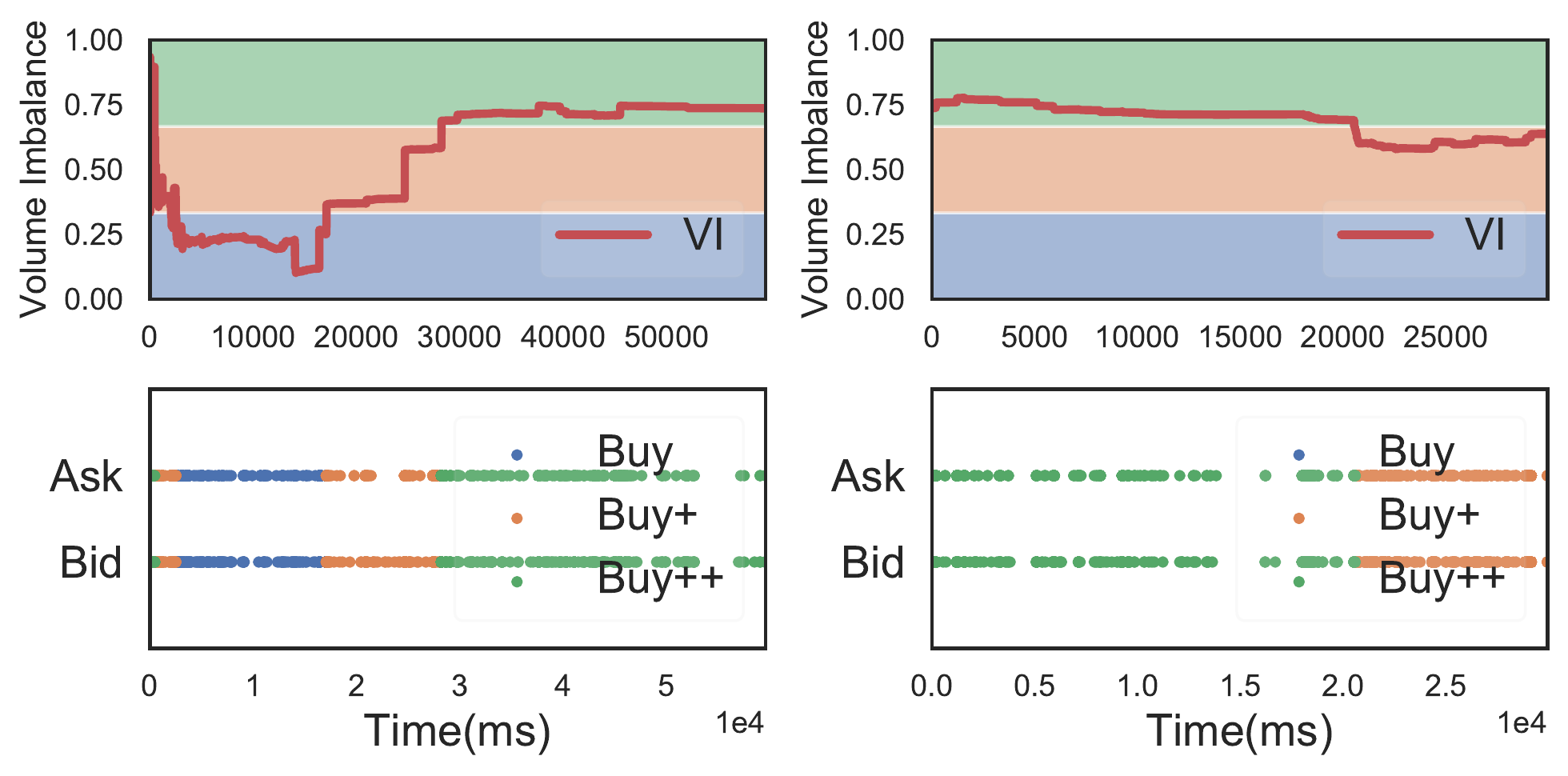}
\subcaption{INTC}\label{app.fig1b}
\end{minipage}
\caption{(a): The timestamps of earthquakes in the SCE training (top) and test (bottom) data. (b): The volume imbalance curve through time and the timestamps of ask and bid orders in the INTC training (left) and test (right) data.} 
\label{app.fig1}
\end{center}
\end{figure}

For the southern California earthquake data, we select earthquakes between latitude 33.8 and 34.8 (north) and between longitude 117.1 and 116.1 (west) with minimum magnitude 3 to guarantee they follow the Gutenberg–Richter law \cite{gutenberg1955magnitude} from January 1, 1976 to December 31, 2020. The data in the same region with a shorter time interval is also used in \citet{wang2012markov}. The timestamps in the training/test data are shown in \cref{app.fig1a}. For the INTC LOB data, the volume imbalance curve and the timestamps of ask and bid orders in the training/test data are plotted in \cref{app.fig1b}. 

We train TR-Hawkes and TM-MRP on the half training sets of SCE and INTC and tune hyperparameters on the another halves as validation sets. We employ the same sets of hyperparameters as described in~\citep{simiao2020transformer} and~\citep{oleksandr2020fast} for tuning TR-Hawkes and TM-MRP, respectively. We note that TR-Hawkes produces an out-of-memory error when training on INTC, so only the first $10000$ events from the training set are used. Also, because the TM-MRP only accepts single dimensional observation, we combine the two dimensional observation in INTC for it. We use a machine with an Nvidia Tesla K40C GPU with 12 GB memory for training these two models. 

For the FS-Hawkes, all hyperparameters are fine tuned. Specifically, for the SCE data, the number of basis functions is set to 12, which are $\tilde{f}_{\{1,\ldots,12\}}=\text{Beta}(\tilde{\alpha}=1,\tilde{\beta}=100, \text{scale}=10, \text{shift}=0)$, $\text{Beta}(\tilde{\alpha}=50,\tilde{\beta}=50, \text{scale}=10, \text{shift}=\{-5,-4,\ldots,4,5\})$. The hyperparameter $\sigma^2$ is set to $1$ by cross validation. The number of quadrature nodes is set to 100 per state interval. The number of iterations is set to 200 for convergence. For the INTC data, the number of basis functions is set to 3, which are $\text{Beta}(\tilde{\alpha}=50,\tilde{\beta}=50, \text{scale}=1, \text{shift}=\{-0.5,-0.4,-0.3\})$. The hyperparameter $\sigma^2$ is set to $1$ by cross validation. The number of quadrature nodes is set to 500 per state interval. The number of iterations is set to 1000 for convergence. For the NL-Hawkes, the basis functions are same as that of FS-Hawkes, other hyperparameters are fine tuned. For the SD-Hawkes, there are no hyperparameters needed to be tuned.  

\begin{figure}[ht]
\begin{center}
\begin{minipage}[b]{0.4\linewidth}
\includegraphics[width=\linewidth]{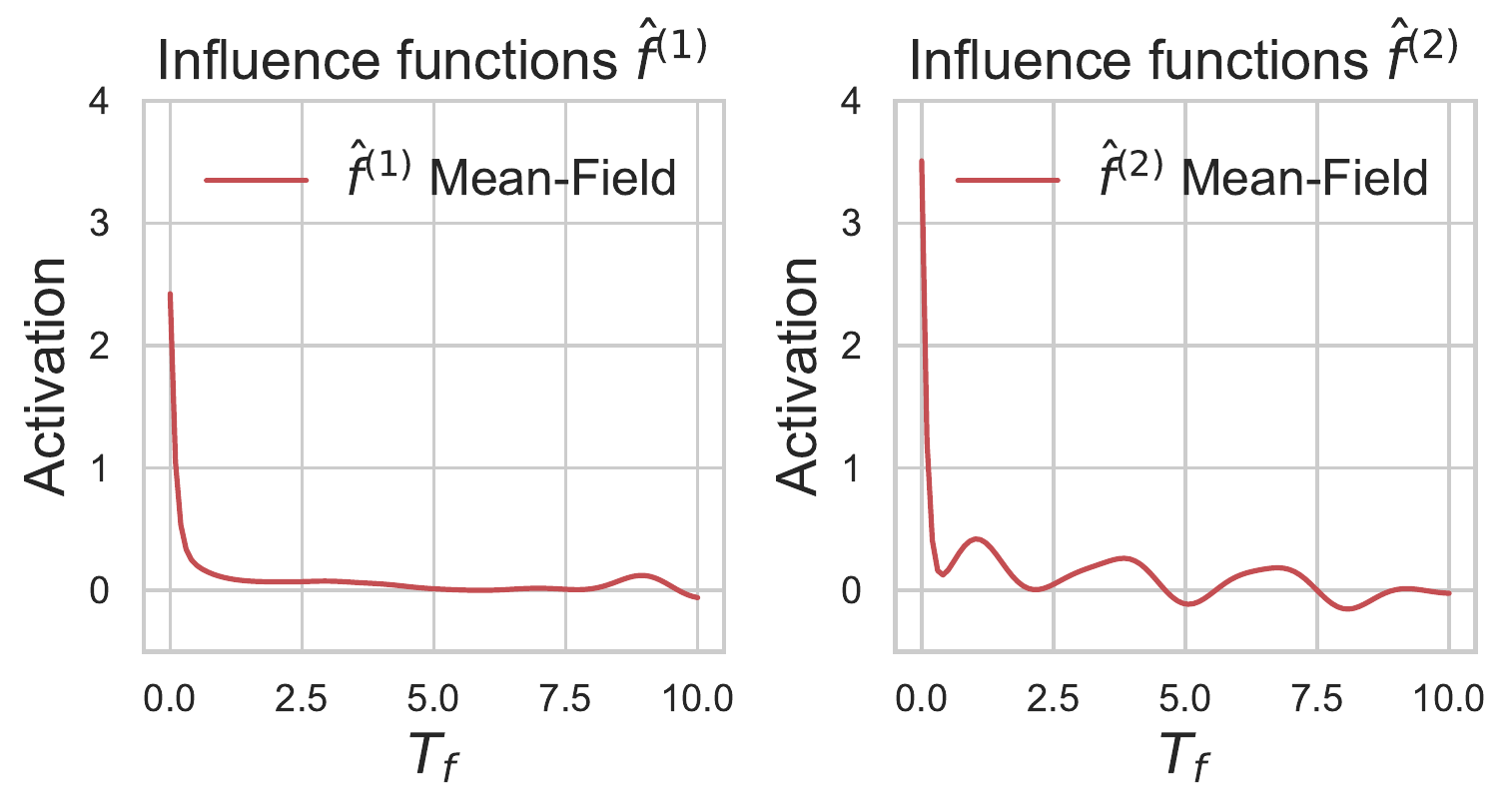}
\subcaption{SCE}\label{app.fig5a}
\end{minipage}
\begin{minipage}[b]{0.7\linewidth}
\includegraphics[width=\linewidth]{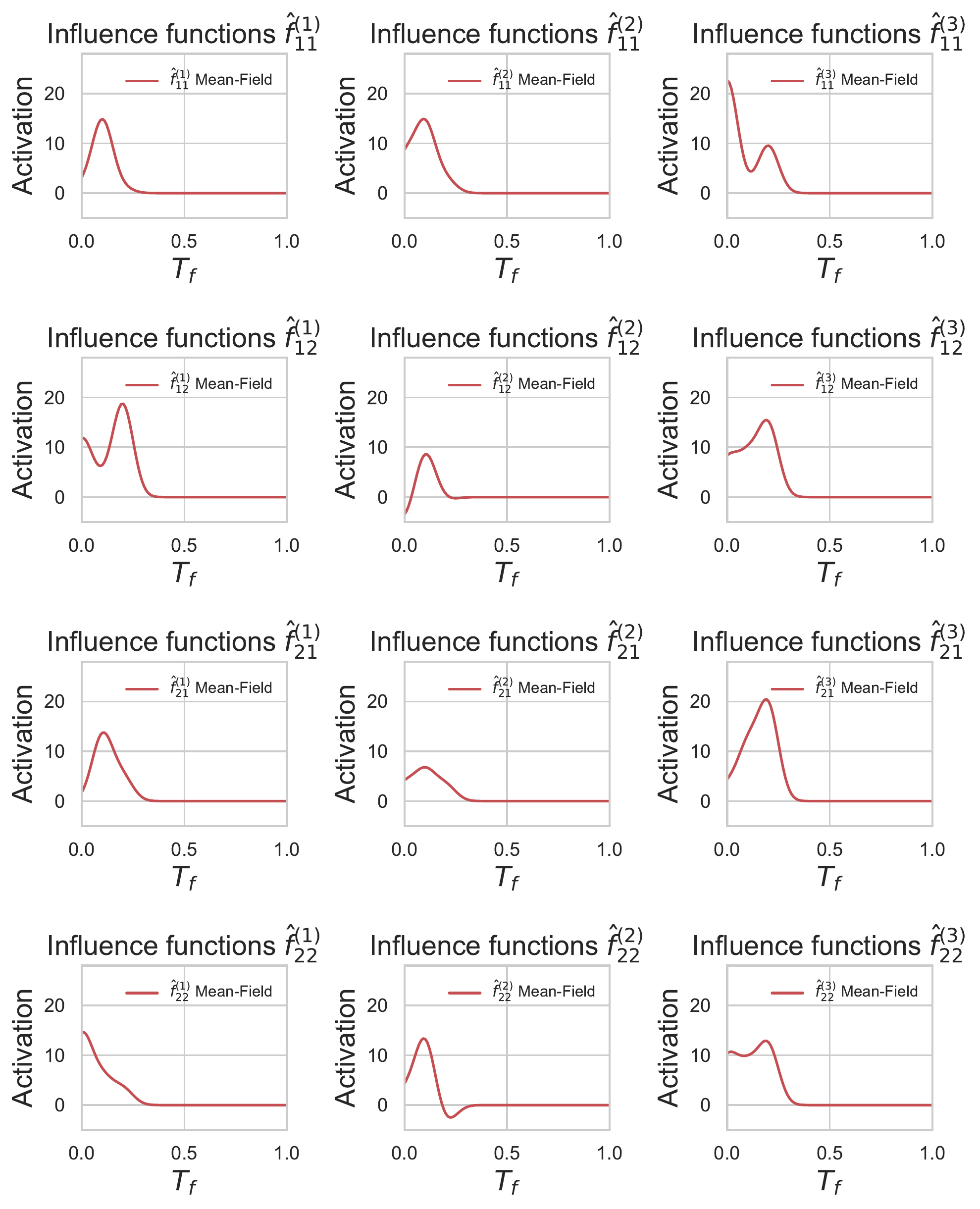}
\subcaption{INTC}\label{app.fig5b}
\end{minipage}
\caption{The estimated influence functions at different states of (a) SCE and (b) INTC.} 
\label{app.fig5}
\end{center}
\end{figure}

The estimated influence functions at different states of SCE and INTC are shown in \cref{app.fig5}. We can see that, no matter which dataset, the influence functions vary a lot at different states, which explains why FS-Hawkes is needed in the time-varying system. 

\end{document}